\documentclass[lettersize,journal]{IEEEtran}
\usepackage{amsmath,amsfonts}
\usepackage{algorithmic}
\usepackage{algorithm}
\usepackage{array}
\usepackage[caption=false,font=normalsize,labelfont=sf,textfont=sf]{subfig}
\usepackage{textcomp}
\usepackage{stfloats}
\usepackage{url}
\usepackage{verbatim}
\usepackage{graphicx}
\usepackage{cite}
\newcommand{\etal}{\textit{et al}.}
\newcommand{\ie}{\textit{i}.\textit{e}.,}
\newcommand{\eg}{\textit{e}.\textit{g}.,}

\usepackage[capitalize]{cleveref}
\crefname{section}{Sec.}{Secs.}
\Crefname{section}{Section}{Sections}
\Crefname{table}{Table}{Tables}
\crefname{table}{Tab.}{Tabs.}

\usepackage{capt-of}

\usepackage{xcolor, colortbl}
\definecolor{LightCyan}{rgb}{0.88,1,1}

\usepackage{pifont}
\newcommand{\xmark}{\text{\ding{55}}}
\newcommand{\cmark}{\text{\ding{51}}}

\usepackage{multirow}

\usepackage[sectionbib]{bibunits} 
\defaultbibliographystyle{IEEEtran}
\defaultbibliography{references}

\begin{document}
\begin{bibunit}

\title{Fast Non-Rigid Radiance Fields\\ from Monocularized Data}

\author{
\hspace{1.6em}Moritz Kappel$^1$\hspace{1.8em}
Vladislav Golyanik$^2$\hspace{1.8em}
Susana Castillo$^1$\hspace{2.6em}\\
Christian Theobalt$^2$\hspace{3.0em}
Marcus Magnor$^1$\vspace{0.8em}\\
{\parbox{\textwidth}{\centering \small $^1$ Computer Graphics Lab, TU Braunschweig, Germany
    \hspace{7pt}{\tt\small \{lastName\}@cg.cs.tu-bs.de}\\
    \small $^2$ Max Planck Institute for Informatics, Saarland Informatics Campus, Germany
    \hspace{7pt}{\tt\small \{lastName\}@mpi-inf.mpg.de}
      }
    }
}

\markboth{\tiny{This work has been submitted to the IEEE for possible publication. Copyright may be transferred without notice, after which this version may no longer be accessible.}}%
{Kappel \MakeLowercase{\textit{et al.}}: Fast Non-Rigid Radiance Fields from Monocularized Data}

\twocolumn[{ 
\renewcommand\twocolumn[1][]{#1} 
\maketitle 
\begin{center} 
    \vspace{-.9cm} 
    \includegraphics[width=0.95\textwidth]{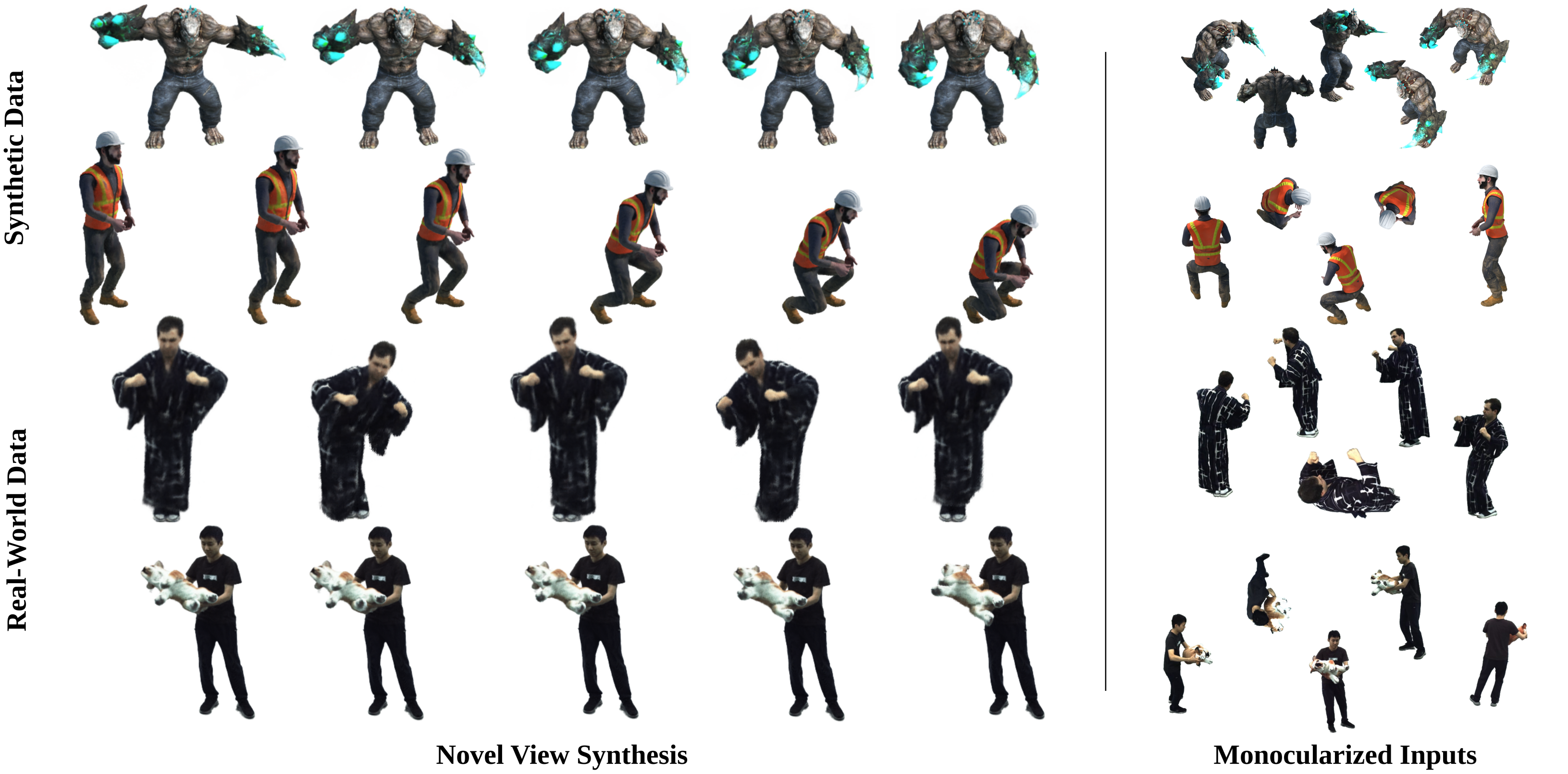} 
    \captionof{figure}{ 
    We introduce a new \textbf{method for real-time volumetric novel view synthesis and temporal interpolation of 360{\textdegree} inward-facing dynamic scenes} (left). 
    It is trained on monocularized sequences, {\ie} only uses a single monocular frame per timestamp randomly sampled from multi-view recordings (right), enabling unprecedented fast optimization at a high visual quality on both synthetic and real-world data.
    } 
    \label{fig:teaser} 
\end{center} 
}] 

\begin{abstract}
    The reconstruction and novel view synthesis of dynamic scenes recently gained increased attention. 
    As reconstruction from large-scale multi-view data involves immense memory and computational requirements, recent benchmark datasets provide collections of single monocular views per timestamp sampled from multiple (virtual) cameras.
    We refer to this form of inputs as \textit{monocularized} data.
    Existing work shows impressive results for synthetic setups and forward-facing real-world data, but is often limited in the training speed and angular range for generating novel views.
    This paper addresses these limitations and proposes a new method for full 360{\textdegree} inward-facing novel view synthesis of non-rigidly deforming scenes. 
    At the core of our method are: 1) An efficient deformation module that decouples the processing of spatial and temporal information for accelerated training and inference; and 2) A static module representing the canonical scene as a fast hash-encoded neural radiance field.
    In addition to existing synthetic monocularized data, we systematically analyze the performance on real-world inward-facing scenes using a newly recorded challenging dataset sampled from a synchronized large-scale multi-view rig.
    In both cases, our method is significantly faster than previous methods, converging in less than 7 minutes and achieving real-time framerates at 1K resolution, while obtaining a higher visual accuracy for generated novel views.  
\end{abstract}



\section{Introduction}\label{sec:intro} 
The faithful reconstruction and rendering of non-rigidly deforming objects from a set of image or video captures is a longstanding challenge for a wide range of practical applications ({\eg} creating  virtual avatars for immersive AR/VR  applications and movie production). 
Recent approaches build upon the success of neural radiance fields~(NeRFs)~\cite{mildenhall2020nerf}, enhancing the scene representation with an additional multilayer perceptron~(MLP) that models temporal deformations in the scene~\cite{pumarola2021d,tretschk2021non,park2021hypernerf, park2021nerfies}.
Similar to NeRF-based methods for static scenes, these non-rigid extensions achieve an unprecedented visual quality for novel view synthesis.
However, the additional temporal dimension not only increases the computational requirements of the system itself, but also drastically increases the amount of training data and thus memory consumption.
As a result, high-quality multi-view reconstruction approaches~\cite{li2021neural, park2021hypernerf} can take several thousand GPU hours to complete, which can quickly become impractical for many scenarios.
The D-NeRF approach by Pumarola~\etal~\cite{pumarola2021d}, on the other hand, introduced a new type of synthetic 360{\textdegree} dataset, where the position of a monocular camera is randomly resampled from a hemisphere for every timestamp.
This idea was later adapted for real-world forward-facing scenes recorded by a stereo camera setup, where the training images are alternatingly sampled from the left and right camera \cite{park2021nerfies, park2021hypernerf}. 
We refer to this input format as monocularized data.
For practical applications, these setups not only enable more efficient training by reducing the storage requirements of conventional multi-view captures, but can also improve the fidelity of dynamic radiance field reconstruction for continuously moving camera trajectories.
In the latter case, the rapidly changing viewing angles of monocularized data largely resolve the problem of occlusion, depth and motion ambiguities which occur for `real' monocular video ({\eg} videos captured by a smartphone) without physically increasing the amount of training data.
Despite the multiple cameras needed to record monocularized sequences, many previous methods falsely advertised the use of such as `monocular' video, as recently criticized by Gao~\etal~\cite{gao2022dynamic}.
Nonetheless, data monocularization entails a huge potential for improving the efficiency when working with large-scale multiview data, enabling, {\eg} the generation of fast interactive previews or operation on systems with memory or compute limitations.
Thus, the synthetic D-NeRF dataset was widely accepted by the community and is now frequently used as a benchmark for new approaches.
However, the applicability to real-world full object reconstruction remains under-explored. 
%

%
In response to the above-mentioned challenges, this paper introduces MoNeRF, a new approach for implicit radiance field reconstruction and novel view synthesis of dynamic scenes from monocularized 360{\textdegree} inward-facing data; see Fig.~\ref{fig:teaser}. 
MoNeRF is general and works for arbitrary non-rigid objects. 
It takes a collection of input images and foreground segmentation masks to reconstruct the depicted scene using a 4D deformation vector field that transforms spatiotemporal samples into a static, canonical radiance field representation.
Our core finding is that a factorization between temporal and spatial domain results in an increased accuracy and allows for efficient training acceleration.
\IEEEpubidadjcol 
Therefore, we propose a fast deformation component that estimates a deformation vector field by splitting temporal and spatial information into individual MLPs.
Furthermore, we adapt the fast hash-encoding proposed by InstantNGP~\cite{mueller2022instant} for representing the scene in a canonical space, making our model several magnitudes faster than previous approaches.
To demonstrate the practical adaptability of monocularized data to real-world inward-facing scenes, we record twelve challenging sequences using a large-scale multi-view setup, and sample a single, static image per timeframe for training. 
We show that, despite the drastic reduction in the number of training views (and thus the compute and memory requirements of full multi-view systems \cite{li2022neural}), our approach is capable of high-fidelity novel view synthesis.
On both synthetic and real data, it yields a superior quality according to several metrics and converges significantly faster than previous methods~\cite{pumarola2021d,tretschk2021non}. 
What is more, the combination of our novel deformation module with a fast hash-encoded scene representation enables real-time framerates of up to 60 FPS during inference -- several orders of magnitude faster than previous approaches -- which makes possible for interactive applications. 
In summary, our main contributions are:  
\begin{itemize} 
    \setlength{\parskip}{2pt} 
    \item A novel method for spatiotemporal novel view synthesis from monocularized sequences on the basis of InstantNGP~\cite{mueller2022instant}, which is trained in under seven minutes and achieves  real-time framerates during inference. 
    \item A neural factorization-based deformation module splitting spatial and temporal information into two MLPs to improve the visual quality and accelerate training. 
    \item A real monocularized dataset with twelve sequences of humans (including loose garments) and general objects, enabling evaluation of fast 360{\textdegree} inward-facing novel view synthesis on dynamic real-world scenes. 
\end{itemize} 
Our dataset and code will be released for research purposes.
\section{Related Work}\label{sec:rw}
\begin{figure*}[ht]
    \centering
    \includegraphics[width=\textwidth,keepaspectratio]{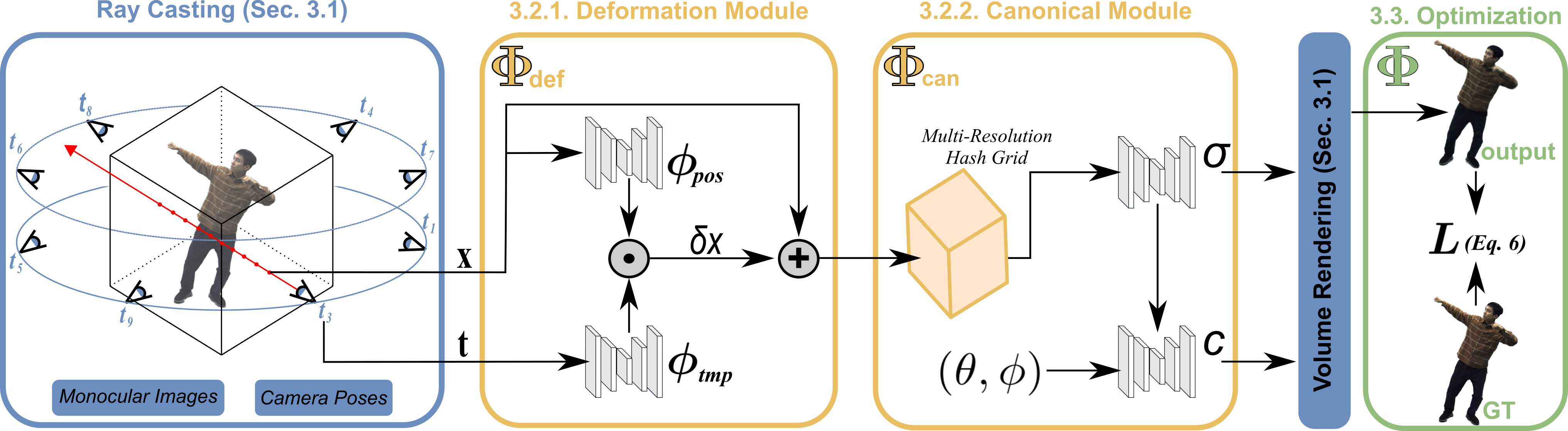}
    \caption{
        \textbf{Framework overview}. Our method takes a set of calibrated monocular RGBA images to reconstruct a deformable radiance field for novel-view synthesis. 
        We feed sampled 3D positions $x$ and their normalized timestamp $t$ into individual shallow MLPs, and combine the resulting high-dimensional embeddings using matrix multiplication to obtain a deformation vector $\delta x$ into canonical space.
        The canonical module is implemented as a fast hash-encoded neural radiance field, estimating opacity $\sigma$ and view-dependent color $c$ for volume rendering.
    } 
    \label{fig:framework} 
\end{figure*} 
The introduction of neural radiance fields (NeRF)~\cite{mildenhall2020nerf} inspired a multitude of follow up works, that adapt and extend it (as well as differentiable volume rendering) for a variety of rendering tasks. 
Prominent extensions propose improvements for anti-aliasing~\cite{barron2021mip}, unstructured in-the-wild data~\cite{martinbrualla2020nerfw, rudnev2022nerfosr}, unbounded scenes~\cite{zhang2020nerf++}, and human avatars~\cite{isik2023humanrf}. 
Our method is mainly concerned with the fast non-rigid scene reconstruction \cite{park2021nerfies, pumarola2021d, li2021neural, tretschk2021non}.
In the following, we briefly review previous work on acceleration and temporal extensions for NeRF. 
A more extensive discussion of neural rendering techniques can be found in the report by Tewari~\etal~\cite{Tewari2022NeuRendSTAR}.
\noindent\textbf{Dynamic Scene Reconstruction and Rendering.} 
The reconstruction and rendering of deforming scenes from a set of input images is a long-standing research area in computer vision and graphics.
Traditional non-rigid reconstructing approaches apply dense structure from motion~\cite{Garg2013,Parashar2020,Sidhu2020,Wang2022CVPR,kumar2022organic}, shape-from-template~\cite{Salzmann2007,Ngo2015,shimada2019ismo,Kairanda2022, johnson2022unbiased}, RGB-D inputs \cite{slavcheva2017killingfusion, bozic2021neural}, or even differentiable physics simulation~\cite{chen2022virtual} to jointly infer information about scene appearance and deformation.
Due to the high visual quality and flexibility of NeRF, a large branch of methods focuses on extensions for non-rigidly deforming scenes, performing \textit{implicit} volumetric radiance field reconstruction for free viewpoint and temporal interpolation.
Full multi-view reconstruction methods use, {\eg} hyper spherical harmonics~\cite{zhang2022neuvv}, Fourier transform~\cite{wang2022fourier}, or compact latent codes~\cite{li2022neural} to optimize the scene on a per frame basis from many input views.
While these approaches achieve an outstanding perceptual quality when rendering novel views, they suffer from immense computational and memory requirements which usually scale with the amount of input data.
On the other hand, some methods add the temporal dimension by modeling the scene as a 4D radiance field, and include regularizers such as motion consistency~\cite{du2021neural, li2021neural}, foreground-background decomposition~\cite{gao2021dynamic, wu2022d} and explicit depth priors~\cite{xian2021space}, to resolve motion and depth ambiguities.
Another family of methods model dynamic deformations as a 4D deformation vector field warping rays into a static (canonical) scene representation.
Previous work further applied explicit rigidity estimation with explicit vector field divergence losses~\cite{tretschk2021non}, per-frame appearance and deformation latent codes~\cite{park2021nerfies}, and hyperplane slicing~\cite{park2021hypernerf} to improve stability.
Our method adopts the split of deformation field and canonical space, but focuses on the fast reconstruction of a full hemisphere (inward-facing scenes). 
Thus, it is most closely related to the D-NeRF approach by Pumarola~\etal~\cite{pumarola2021d}, who introduced a synthetic monocularized dataset to reconstruct dynamic objects.
Our method, however, significantly accelerates training and inference, while simultaneously improving the visual fidelity through a new factorization-based deformation module coupled with an explicit static scene representation.
Furthermore, we demonstrate the practicability and efficiency of our approach on real-world multi-view captures using a new dataset generated by monocularizing ({\ie} discarding all but one view per timeframe) inward-facing recordings from a large multi-view camera rig.
In contrast to full multi-view reconstruction, our method and setup enable reconstruction in a matter of minutes due to the immense reduction in compute and memory requirements, and achieve real-time framerates during inference.
\noindent\textbf{NeRF Acceleration.}
The training and rendering speed of conventional NeRF methods is often limited by the amount of ray-marching samples and expensive network queries during volume rendering. 
One branch of research focuses on improving inference times by applying efficient acceleration structures to the radiance field~\cite{garbin2021fastnerf, hedman2021baking, reiser2021kilonerf, yu2021plenoctrees}, or introducing advanced ray sampling and stopping criteria~\cite{lindell2021autoint, piala2021terminerf}.
Another branch attempts to accelerate training using generalized pretrained models to fit novel scenes from one or few input views~\cite{yu2021pixelnerf, chen2021mvsnerf, liu2022neural}.
Recently, multiple works introduced explicit scene representations such as  discrete grids~\cite{sun2022direct, yu2021plenoxels}, to replace the costly MLP queries with efficient data lookups and trilinear interpolation. 
However, the reduced computational cost of dense data grids usually implies an increase in memory requirement, which can limit the obtainable rendering quality.
To reduce the increased memory footprint, Chen~\etal~\cite{Chen2022ECCV} model the explicit volume as a 4D tensor factorized into compact low-rank tensor components.
Müller~\etal~\cite{mueller2022instant} on the other hand propose an explicit feature encoding for MLP-based graphics primitives (including NeRFs) using multi-resolution hash grids.
They work remarkably well on rigid scenes, but their adaptability for non-rigid extensions remains unclear. 
This work builds upon the success of Instant-NGP~\cite{mueller2022instant}, introducing an efficient temporal extension that leverages hash encoding for non-rigid reconstruction.
\noindent\textbf{Fast Non-Rigid Radiance Fields.}
Very recently, other non-rigid NeRF methods adapt explicit scene representations to improve the training speed in various scenarios.
Guo~\etal~\cite{guo2022neural} adapt efficient direct voxel grid optimization~\cite{sun2022direct} to model scene deformation, density and color.
Along the same lines, Fang~\etal~\cite{fang2022fast} use a time-aware neural voxel grid with multi-distance interpolation for fast, high-quality non-rigid reconstruction.
Another class of methods~\cite{kplanes_2023, shao2023tensor4d, cao2023hexplane} extend the factorization approach of TensoRF~\cite{Chen2022ECCV} to model additional dimensions such as time.
MixVoxels~\cite{wang2022mixed} further enhances their model by distinguishing between static and dynamic voxels to improve the modeling of background and scene motion.
NeRFPlayer~\cite{song2023nerfplayer} introduces a spatiotemporal decomposition and feature streaming scheme to reconstruct dynamic scenes from hand-held cameras in up to 10 seconds per frame.
HyperReel~\cite{attal2023hyperreel} proposes a ray-conditioned sample prediction network combined with a keyframe approach for fast 6-DoF video rendering.
The recent NerfAcc toolbox~\cite{li2023nerfacc} gathers acceleration strategies from recent literature, and combines them to an easy plug-and-play PyTorch framework, including extensions for dynamic scenes.
Many of these approaches achieve an impressive performance on the synthetic monocularized D-NeRF~\cite{pumarola2021d} dataset, but do not evaluate the performance on real-world 360{\textdegree} inward-facing recordings.
We show that in both scenarios, our hash-grid-based method trains faster while obtaining higher average scores than related benchmark approaches according to image quality metrics, and achieves higher ({\ie} real-time) framerates for novel-view rendering.

\section{Method}\label{sec:method}
Given a set of $N$ calibrated input images $I = (i_1, ..., i_N)$ depicting a non-rigidly deforming object, the corresponding foreground masks $S = (s_1, ..., s_N)$, and per-frame normalized timestamps $T = (t_1, ..., t_N) \in [0, 1]^{N}$, our model enables joint novel view synthesis and seamless temporal interpolation in real time. 
As illustrated in \cref{fig:framework}, the core of our model comprises an explicit hashgrid-based radiance field representing the scene in canonical ({\ie} static undeformed) space, which is optimized using conventional neural volume rendering.
We further incorporate temporal deformations in our model by prepending an efficient ray-bending component transforming spatiotemporal samples into canonical space.
In the following, we provide a detailed description of our individual framework components and the optimization procedure. 

\subsection{Volume Rendering}
To optimize and render novel views from our model, we apply discrete sample-based ray marching derived from classical volume rendering \cite{kajiya1984ray} as introduced 
in the context of NeRF \cite{mildenhall2020nerf}. 
As these fundamental concepts have been extensively discussed in the  aforementioned literature, we provide a brief introduction of the  mathematical concepts and notation necessary to delineate our method. 
To query the expected color of a single pixel $p$ in a camera view $n$, we first construct a ray $r(p) = o + pd$ starting at the camera's optical center $o$ and passing through the pixel's center in direction $d$. 
For any continuous scene function $\boldsymbol{\Phi}$ ({\eg} NeRF) mapping a spatial position $x \in \mathbb{R}^{3}$ and viewing direction $d \in \mathbb{R}^{3}$ to a density $\sigma \in \mathbb{R}$ and color $c \in \mathbb{R}^{3}$, \textit{i.e.,} 
\begin{equation} \label{eq:model_static}
(\sigma, c) = \boldsymbol{\Phi}(x, d),
\end{equation}
we can then evaluate the local scene properties $(\sigma_{i}, c_{i}), i \in \{1, ..., M\}$ for a set of $M$ discrete spatial samples $x_{i} = (r(p_{i}), p_{i} \sim [p_{n}, p_{f}])$ along the ray within a predefined minimal ($p_{n}$) and maximal ($p_{f}$) distance from the image plane. 
The final pixel color is obtained by estimating the integral $C(r)$ over all samples according to the optical model of Max~\cite{max1995optical}: 
\begin{align} \label{eq:color_rendering} 
C(r) = \sum_{i=1}^{M} T_{i}(1 - \textrm{exp}(-\sigma_{i}\delta_{i}))c_{i}, 
\end{align} 
where $T_{i} = \textrm{exp}(-\sum_{j=1}^{i-1}\sigma_{j}\delta_{j})$ and  $\delta_{i} = p_{i+1} - p_{i}$ denotes the distance between adjacent  samples. 
Finally, we use the pixel's total transmittance $\alpha(r)$ given by 
\begin{equation} 
\alpha(r) = \sum_{i=1}^{M} T_{i}(1 - \textrm{exp}(-\sigma_{i}\delta_{i})) 
\end{equation}
to blend the estimated color with a static background color. 
The following sections describe how our model represents the local scene deformation, density and color for rendering non-rigid scenes. 
\subsection{Scene Representation}
The scene function \eqref{eq:model_static} has no notion of time, and thus cannot reflect objects undergoing non-rigid deformations. 
Inspired by previous work on deforming radiance fields \cite{pumarola2021d, tretschk2021non}, we introduce a framework consisting of two distinct subcomponents to handle the temporal dimension: A static module $\boldsymbol{\Phi}_{\textrm{can}}$ representing the scene in canonical ({\ie} undeformed) space, and a deformation module $\boldsymbol{\Phi}_{\textrm{def}}$ estimating the offset of a spatial point at timestamp $t$ into its canonical state.
Formally, the full scene function implemented in our framework can be rewritten as:
\begin{equation} \label{eq:model_dynamic}
    \boldsymbol{\Phi}(x, d, t) = \boldsymbol{\Phi}_{\textrm{can}}(x + \boldsymbol{\Phi}_{\textrm{def}}(x, t), d). 
\end{equation}

\subsubsection{Deformation Module}\label{sec:deformation_module}
The deformation module $\boldsymbol{\Phi}_{\textrm{def}}$ estimates the deformation vector $\delta x \in \mathbb{R}^{3}$ for a discrete spatial position $x$ at a normalized timestamp $t$. 
This process was previously described as a form of ray bending, where rays cast by a virtual video camera are distorted so that related samples along the rays can be evaluated at their original location in a canonical volume~\cite{tretschk2021non}. 
Most recent methods implement ray bending using a single multilayer perceptron (MLP) of a similar size as the canonical NeRF itself~\cite{pumarola2021d, tretschk2021non, park2021nerfies}. 
While this approach can represent high-quality deformation fields, it also significantly contributes to the overall computational cost, both during training and inference.
In contrast, we propose a new deformation module architecture decoupling the spatial and temporal dimensions into individual MLPs, which significantly reduces training and inference times of our model while preserving a high quality:
\begin{equation}
\delta x = \boldsymbol{\Phi}_{\textrm{def}}(x, t) = \boldsymbol{\phi}_{pos}(x) \cdot \boldsymbol{\phi}_{tmp}(t), 
\end{equation}
with $\boldsymbol{\phi}_{pos}:\mathbb{R}^{3} \rightarrow \mathbb{R}^{3 \times l}$ denoting the positional MLP, $\boldsymbol{\phi}_{tmp} : \mathbb{R} \rightarrow \mathbb{R}^{l \times 1}$ denoting the temporal MLP, and $\cdot$ being conventional matrix multiplication.
Similar to recent scene factorization approaches~\cite{Chen2022ECCV, kplanes_2023}, the key idea of our approach is to achieve an explicit disentanglement of our deformation vector field using separate temporal and spatial embeddings.
To this end, instead of directly estimating the per-point offsets $\delta x$ in a single MLP from a concatenation of temporal and spatial information, we first estimate a higher-dimensional feature vector of size $l$ for every spatial input dimension.
The resulting spatial embedding is then reduced to a 3D deformation vector by linear combination with temporal embedding coefficients estimated by the second MLP. 
With sufficient capacity, {\ie} a large enough $l$, this representation can support arbitrary deformation fields and does not limit the overall model expressivity.
In combination with our hash-encoded canonical module, this procedure offers several advantages over single MLP-based deformation modelling:
As both networks produce embeddings for individual inputs with different semantics, they enable varying implicit regularization for temporal and spatial information.
We further encourage this behaviour by applying different types of input encodings.
More precisely, we use frequency encoding~\cite{mildenhall2020nerf} for the spatial input position, which enables higher frequencies for cleaner distinction between static and dynamic scene contents.
For the input timestamp $t$, we leverage one-blob encoding~\cite{muller2019neural} to obtain a smooth temporal embedding, which helps to reduce temporal jittering when training on monocularized sequences like the D-NeRF dataset~\cite{pumarola2021d}. 
Furthermore, our deformation field factorization policy enables the use of more shallow network architectures with overall fewer input parameters; 
it accelerates training while maintaining an overall high quality of the deformation field. 
This speed up is particularly noticeable when querying deformation vectors for the same spatial position at multiple points in time, as the positional embedding can be efficiently precomputed and later combined with different temporal embeddings in parallel.
While, at first glance, this property does not benefit the ray-marching-based rendering, it significantly speeds up the update of our efficient acceleration structure described in \cref{subsec: acceleration}, leading to drastic improvements in training and inference speed.
\subsubsection{Canonical Module}
After shifting the input samples to the canonical space by adding the estimated offset $\delta x$, we apply a static radiance field representation $\boldsymbol{\Phi}_{\textrm{can}}$ to obtain their local density $\sigma$ and view-dependent color values $c$ as stated earlier in \cref{eq:model_static}.
For this purpose, we adapt the recent InstantNGP introduced by Müller~\etal~\cite{mueller2022instant}, which reconstructs detailed radiance fields in a few minutes at a competitive visual quality. 
The authors achieve this unprecedented level of speed by replacing the frequency-based input encoding~\cite{mildenhall2020nerf} with an explicit multi-resolution hash grid encoding. 
From this grid, learned feature vectors can be queried in constant time using trilinear interpolation, which are processed by two shallow fully-fused MLPs to obtain the desired output values.
For our framework, we employ a fast PyTorch~\cite{paszke2019pytorch} re-implementation using the architectural hyperparameters provided in the original publication. 
\subsection{Optimization}
We jointly optimize our full model $\boldsymbol{\Phi}$ \eqref{eq:model_dynamic} end-to-end over the course of $30$K training iterations via stochastic gradient descent.
During each iteration, we chose a random camera view $n \in \{1, ..., N\}$ with associated image $i_n$, foreground mask $s_n$, and timestamp $t_n$. 
For this view, we uniformly sample a batch $R$ of $8192$ rays, which are rendered using a maximum of $512$ samples per ray.
While it is theoretically possible to sample a batch of rays from all available training views, this single image sampling approach speeds up the optimization, as only a single timestamp $t_n$ needs to be evaluated by the temporal MLP $\boldsymbol{\phi}_{tmp}$.
Our full objective function $\boldsymbol{L}$ consists of three terms:
\begin{equation}
    \boldsymbol{L} = \boldsymbol{L}_{\textrm{photo}} + \lambda_{\textrm{bg}}\boldsymbol{L}_{\textrm{bg}} + \lambda_{\textrm{def}}\boldsymbol{L}_{\textrm{def}}, 
\end{equation}
with scalar hyperparameters $\lambda_{\textrm{bg}}$ and $\lambda_{\textrm{def}}$ that we set to $10^{-2}$ and $10^{-3}$ for all our experiments, respectively. 
The photometric loss $\boldsymbol{L}_{\textrm{photo}}$ compares the estimated pixel color $c_{\textrm{estim}}$ to the ground truth pixel color $c_{\textrm{gt}}$: 
\begin{equation} 
    \boldsymbol{L}_{\textrm{photo}} = \dfrac{1}{|R|} \sum_{r \in R}||C(r) - C_{\textrm{gt}}(r)||_{2}^{2}, 
\end{equation}
where $C_{\textrm{gt}}(r)$ denotes the ground truth color of ray $r$ in the training view $i_n$.
Following Müller~\etal~\cite{mueller2022instant}, we stabilize training by applying a random background color during each iteration based on the ray transmittance $\alpha (r)$ and foreground mask $s_n$ for the estimated and ground-truth color respectively.
We further apply two regularization losses to stabilize the optimization and improve the generalization to novel views. 
The background entropy loss $\boldsymbol{L}_{\textrm{bg}}$ enforces a clear transition between foreground object and empty scene space, while the deformation field regularizer $\boldsymbol{L}_{\textrm{def}}$ encourages the deformations to be small and sparse:
\begin{align}
    \boldsymbol{L}_{\textrm{bg}} =& \dfrac{1}{|R|} \sum_{r \in R}-\alpha_{r}\log (\alpha_{r}),\label{eq:lbg}\\
    \boldsymbol{L}_{\textrm{def}} =& \dfrac{1}{|X_{\delta}|} \sum_{\delta x \in X_{\delta}}||\delta x||_{1}.\label{eq:ldef}
\end{align}
Here, $X_{\delta}$ denotes the set of all deformation vectors estimated via \cref{eq:model_dynamic} during rendering.
The entire optimization takes 6-7 minutes on a single NVIDIA RTX 3090 GPU.
More implementation details can be found in our supplemental material and official source code release, which will be publicly available. 
\subsection{Acceleration Strategies} \label{subsec: acceleration}
\begin{figure}[bth]
    \centering
    \includegraphics[width=\linewidth, keepaspectratio]{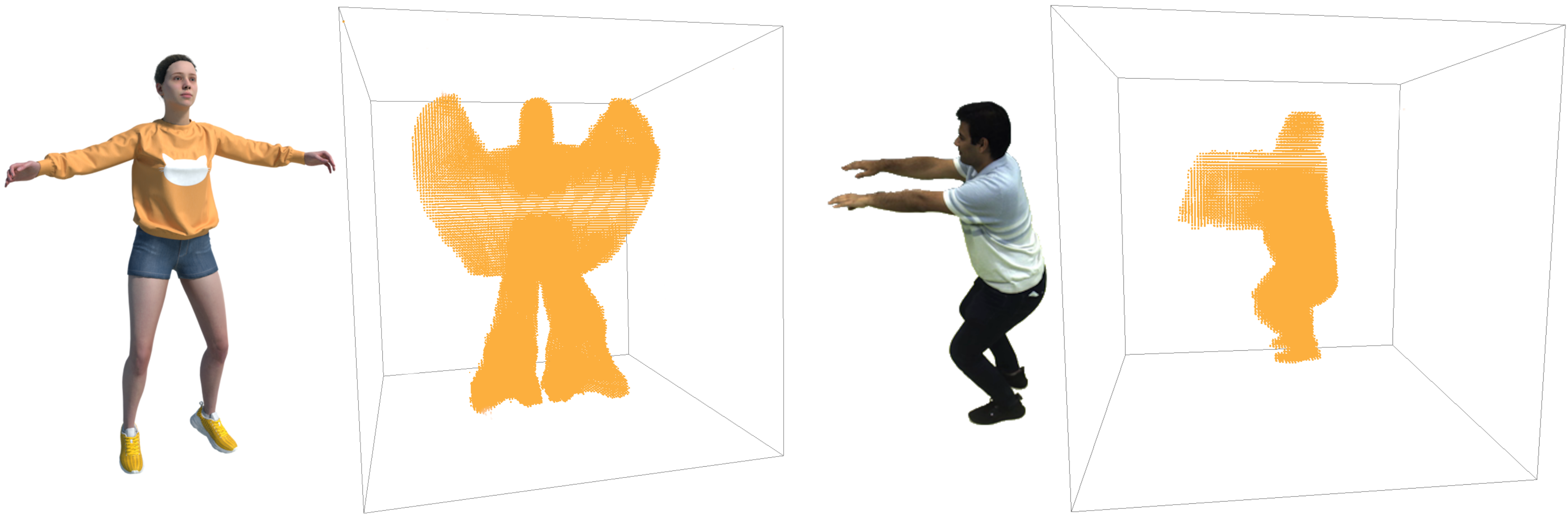}
    \caption{
        \textbf{A point cloud visualization of our temporal occupancy grids.} Left: A synthetic human avatar performing jumping jacks. Right: A real recording of an actor performing squats.
} 
    \label{fig:occupancy} 
\end{figure}

An advantage of our method is that---despite adding a temporal extension to the radiance field---we can directly apply the ray marching acceleration techniques proposed for instant NeRF, such as transmittance-based stopping criteria to speed up training and inference times~\cite{mueller2022instant}. 
Another common way to increase inference speed is keeping track of an occupancy grid, which marks unoccupied (empty) scene space that can be skipped during ray marching~\cite{mueller2022instant, sun2022direct}.
However, naively applying an occupancy grid to our canonical module implies that the deformation module needs to be executed for all samples before being able to skip single evaluations in canonical space, which would result in a significant performance loss. 
Thus, similar to the concurrent NerfAcc~\cite{li2023nerfacc} toolbox, we extend this acceleration structure in the form of a temporal occupancy grid, which marks points in space that are occupied during any timestamp in the normalized time period $[0, 1]$, as visualized in \cref{fig:occupancy}. 
This way, both the canonical and deformation module are only evaluated on a sparse subset of the scene to model the foreground object and resolve the (dis)occluded areas. 
We can then update our temporal grid by sampling a set of candidate cells as proposed by Müller~\etal~\cite{mueller2022instant} for the rigid case, and threshold the accumulated density from a set of $l$ random equidistant timestamps. 
For all our experiments, we use $q=20$ temporal samples, which, in contrast to the naive NerfAcc implementation, can efficiently be evaluated in parallel by our novel deformation model, enabling faster and more frequent updates.
\section{Monocularized Multi-View Avatars Dataset}\label{sec:dataset} 
After its release, despite the monocularized nature of the generation setup, the synthetic D-NeRF~\cite{pumarola2021d} dataset quickly became a common benchmark for new dynamic radiance field approaches.
As the rapidly changing viewing angles of the `teleporting' camera observe adjacent deformation states from varying perspectives, this dataset enables high-quality full 360{\textdegree} novel view synthesis over the entire temporal sequence.
In contrast to large-scale multi-view dataset like the recently released ActorsHQ~\cite{isik2023humanrf} and ZJU-MoCap~\cite{peng2021neural} datasets, which can be used to reconstruct individual high-quality radiance fields on a per-frame basis, this monocularized data is compact in size and thus drastically reduces the compute and memory requirement of dynamic radiance field training.
This idea was later adapted for real-world forward-facing stereo recordings~\cite{park2021hypernerf}, enabling efficient training and testing by leaking multi-view information into the single-view training sequence.
Especially in the case of large-scale multi-view recordings, data monocularization can become useful to enable, {\eg} the generation of fast previews to validate the expected quality of camera recordings and calibration, or enable training on systems with natural memory, compute or network limitations.
However, despite the popularity of the synthetic D-NeRF dataset, this setup and corresponding methods were not yet evaluated in the context of real-world inward-facing full 360{\textdegree} captures, which can allow to reconstruct single dynamic objects for various multimedia applications.
To investigate the applicability of the monocularized data setup for real-world inward-facing scenes, we record new sequences in the spirit of the synthetic D-NeRF dataset.
We dub our novel dataset Monocularized Multi-View Avatars (MMVA).
It comprises a total of $12$ sequences of actors performing motion in clothing of varying complexity or interacting with general non-rigidly deforming objects, captured by a large inward-facing multi-view setup of $m = 109$ synchronized $50$Hz cameras.
In contrast to recent multi-view datasets that specialize exclusively on human motion and related tasks like novel pose synthesis~\cite{isik2023humanrf, peng2021neural}, these complex clothes and object interactions are designed as a challenge for general, unconstrained reconstruction methods.
Each sequence of our dataset consists of $m$ RGB videos with a length of $n \in [100, 250]$ frames at a $1028 \times 752$ pixels resolution, and the corresponding foreground masks obtained via background subtraction.
We then monocularize multi-view recordings as follows:
First of all, we reserve two expressive static camera views for validation and testing.
Then, for every temporal frame $t \in (1, ..., n)$ in the recording sequence, we uniformly sample a single camera index $c \in (1, ..., m)$ from the remaining views, extract the corresponding image at time $t$ from chosen camera $c$, and concatenate the segmentation mask to obtain a single RGBA image.
Thus, the resulting $12$ sequences consist of only $n$ monocular training views with temporally varying camera positions and viewing directions.
In contrast to the full multi-view videos which take several dozen GB in size and require live-decoded during training, the monocularized sequences are only ${\sim500}$ MB in size (including the validation and test data), which can be efficiently preloaded while maintaining an effective multi-view signal over time.
We show that, despite the unique challenges of real world data, such as inaccuracies in intrinsic and extrinsic camera calibration, synchronisation, as well as complex scene illumination and surface reflections, MoNeRF can exploit data monocularization to perform fast NeRF reconstruction in minutes, while maintaining high visual quality for novel view synthesis. 
In the future, our dataset can help to further investigate the robustness of dynamic radiance field approaches on real-world data. 
To this end, our dataset will be made publicly available.
\section{Experimental Evaluation}\label{sec:experiments} 
We perform extensive experiments on an established synthetic dataset and our newly recorded real-world MMVA sequences. 
For quantitative evaluation on the D-NeRF dataset, we gather the reported results from recent literature.
We further conduct more extensive qualitative comparisons to relevant state-of-the-art (SOTA) methods related to our method.
More precisely, we compare our method to D-NeRF~\cite{pumarola2021d} and NR-NeRF~\cite{tretschk2021non},
which both combine a deformation vector field with an implicit canonical scene representation, and the recent TiNeuVox~\cite{fang2022fast}, which is the current state-of-the art in terms of training time and image quality.
For completeness, we also report results for InstantNGP~\cite{mueller2022instant}, which uses the same static scene representation as MoNeRF, but without any temporal component. 
We report three metrics for quantitative assessment: Peak signal-to-noise ratio (PSNR), structural  similarity~(SSIM)~\cite{wang2004image}, and learned perceptual image patch similarity~(LPIPS)~\cite{zhang2018unreasonable}. 
While the PSNR reflects per-pixel the error and is thus closest to the training objective function, SSIM and LPIPS gauge the perceptual reconstruction accuracy from a larger context. 
Video results and additional per scene visualizations and metrics are available in our supplemental material. 
\subsection{Synthetic Scenes} 
\begin{figure*}[ht]
    \centering
    \includegraphics[width=\textwidth,keepaspectratio]{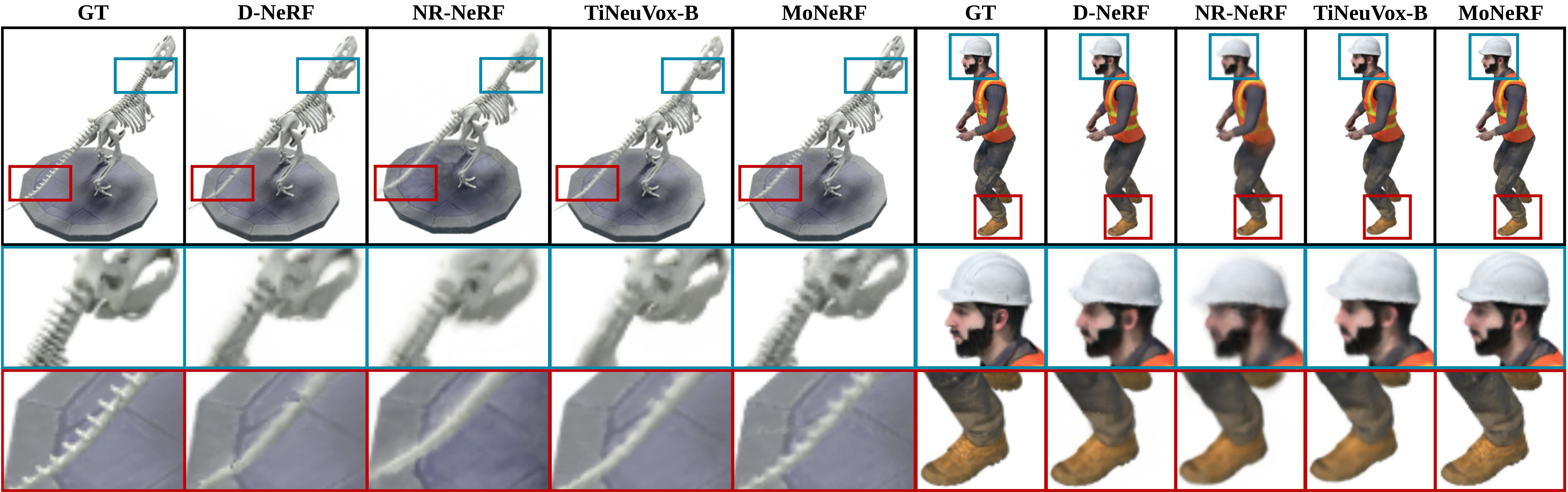}
    \caption{
        \textbf{Comparison on the synthetic D-NeRF dataset~\cite{pumarola2021d}}. We show detailed comparisons of our results (MoNeRF) against ground-truth (GT) and benchmark methods for the \textit{T-Rex} and \textit{Stand Up} sequences.
    } 
    \label{fig:CompDNeRF} 
\end{figure*}

\begin{table}[b!]
    \caption{\textbf{Quantitative comparison on the D-NeRF Dataset~\cite{pumarola2021d}}. We report image quality metrics and training times averaged over all eight sequences. MoNeRF training times were measured on a single NVIDIA RTX 3090 GPU. Values for all other methods are taken from the original papers when available. 
    }\label{tab:AvgMetricsDNeRF}
    \centering
    \footnotesize{
    \setlength\tabcolsep{4.5pt}
    \begin{tabular}{c|l|ccc|c|}
        &\multirow{2}{*}{\textbf{Method}} & \multicolumn{4}{c|}{\textbf{D-NeRF Dataset}}\\
        & & \textbf{PSNR $\uparrow$} & \textbf{SSIM $\uparrow$} & \textbf{LPIPS $\downarrow$} & \textbf{Time}\\ 
        \hline
        \vspace{-8pt}
        \parbox[t]{2mm}{\multirow{4}{*}{\rotatebox[origin=c]{90}{STATIC}}} & & &  &  &\\      
        &NeRF~\cite{mildenhall2020nerf} & 19.00 & 0.87 & 0.18 & $\sim$ h\\
        &DirectVoxGO~\cite{sun2022direct} & 18.61 & 0.85 & 0.17 & 5 min \\
        &Plenoxels~\cite{yu2021plenoxels}  & 20.24 & 0.87 & 0.16 & 6 min\\
        &InstantNGP~\cite{mueller2022instant} & 19.00 & 0.88 & 0.17 & 5 min\\
        \hline
        \vspace{-8pt}
        \parbox[t]{2mm}{\multirow{8}{*}{\rotatebox[origin=c]{90}{DYNAMIC}}} &  &  &  & &\\      
        &T-NeRF~\cite{pumarola2021d} & 28.78 & 0.95 & 0.07 & $\sim$ h\\
        &D-NeRF~\cite{pumarola2021d} & 29.67 & 0.95 & 0.06 & $\sim$ h\\
        &NR-NeRF~\cite{tretschk2021non} & 26.15 & 0.95 & 0.09 & $\sim$ h\\ 
        &TiNeuVox-S~\cite{fang2022fast} & 30.75 & 0.96 & 0.07 & 8 min\\
        &TiNeuVox-B~\cite{fang2022fast} & \textbf{32.67} & 0.97 & 0.04 & 28 min\\
        &NDVG \cite{guo2022neural} & 31.32 & 0.97  & 0.05 & 34.78 min\\ 
        &$K$-Planes-hybrid~\cite{kplanes_2023} & 31.61 & 0.97  & -- & 52 min\\
        \cline{2-6}
        & MoNeRF (ours) & \textit{32.16}  & \textbf{0.98} & \textbf{0.03} & 7 min
    \end{tabular}
    }
\end{table}

\begin{figure}[ht]
    \centering
    \includegraphics[width=1.0\linewidth, keepaspectratio]{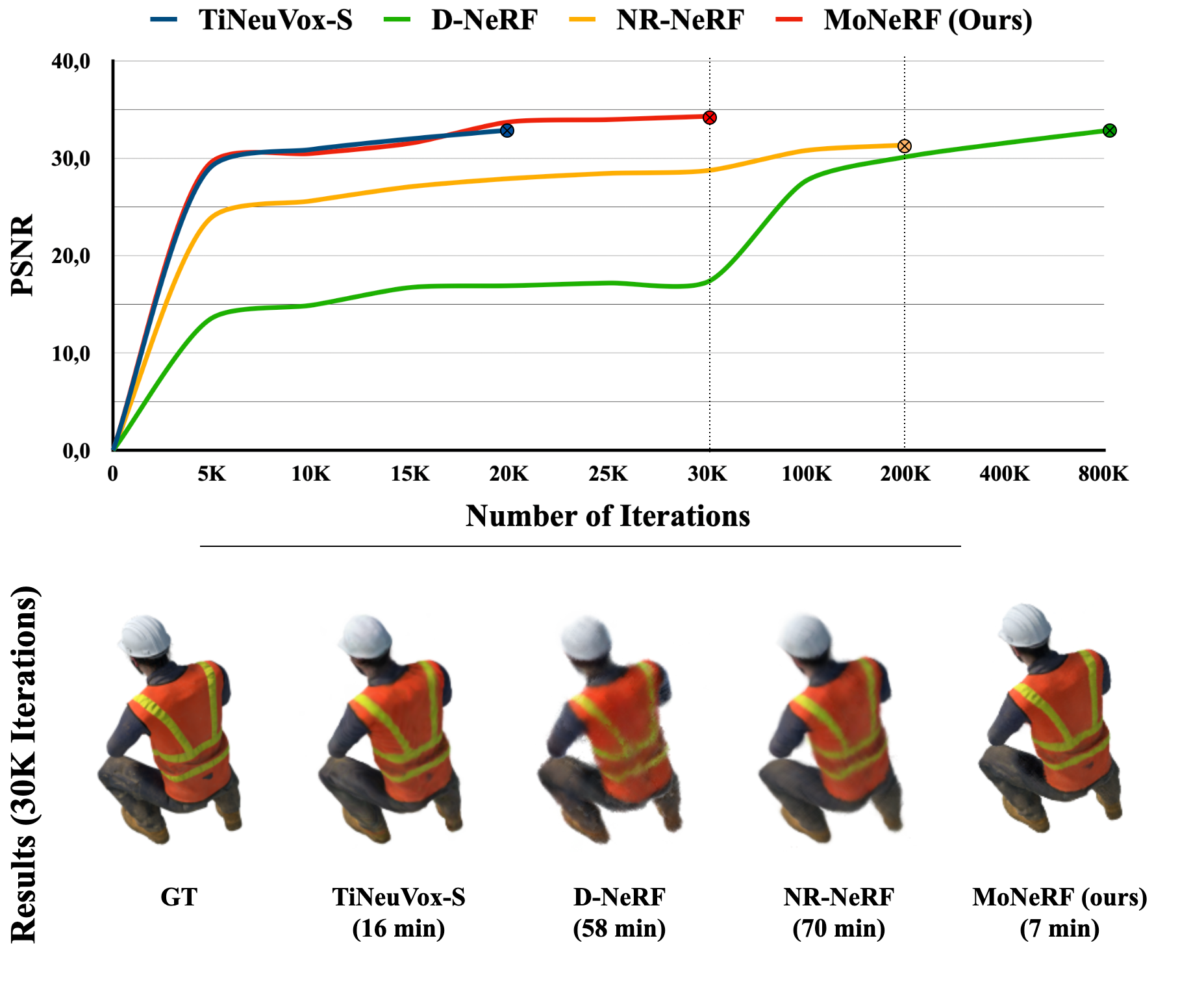}
    \caption{%
        \textbf{Training convergence} of different methods on the synthetic \textit{Stand Up} scene. 
        (Top): A plot showing each method's average test set PSNR for the individual default number of iterations.
        (Bottom): An exemplary test frame and the corresponding training time after $30$K iterations.
        At this stage, compared to recent approaches, MoNeRF produces the most detailed results in only a fraction of the training time. 
    } 
    \label{fig:convergence} 
\end{figure}

\begin{table}[ht]
    \caption{\textbf{Average time consumption} of individual framework components on the D-NeRF sequences. Dataset loading includes training, test and validation sets. Output generation includes model checkpoint, testset rendering, and quality metric calculation for all 20 test frames.}
    \label{tab:timings}
    \centering
    \footnotesize{
    \setlength\tabcolsep{3pt}
    \begin{tabular}{l|c|c|c}
        \textbf{Task} & \textbf{Iterations} & \textbf{Time / Iteration [ms]} & \textbf{Total Time [s]} \\
        \hline
        Load Dataset & 1 & 8435 & 8.5 \\
        Initialize Model & 1 & 566 & 0.5 \\
        Update Model & 30000 & 8 & 238 \\
        Update Grid & 1875 & 78 & 146 \\
        Write Outputs & 1 & 8253 & 8 \\
        \hline
        \multicolumn{3}{c|}{} & \textbf{6.7 min}\\
    \end{tabular}
    }
\end{table}
We use the D-NeRF dataset~\cite{pumarola2021d} to test and compare our method on synthetic data.
It comprises eight sequences, each consisting of $50{-}200$ frames, showing deforming objects of varying motion complexity. 
Similar to previous work, we downsample the images to half resolution ($400\times400$ pixels) for fair comparison.
\cref{tab:AvgMetricsDNeRF} shows average quantitative results over all eight sequences, including the average training time as reported in the original papers. 
Compared to previous dynamic NeRF approaches with implicit canonical radiance fields~\cite{pumarola2021d, tretschk2021non} that take approximately 20 hours to reconstruct a single scene, our novel deformation module and discrete hash-encoded scene representation achieve significant improvements in the training speed (finishing training in under seven minutes and rendering novel views in real time). 
A detailed list showing the time usage of each method component can be found in \cref{tab:timings}. As indicated by the high computational cost per iteration, maintaining our temporal occupancy grid is one of the most expensive tasks, but in return enables extremely efficient model updates, which are executed more frequently.
The quality metrics show that our approach outperforms these methods in terms of image quality, maintaining fine details in high-frequency areas, as shown in \cref{fig:CompDNeRF}.
The state-of-the-art TiNeuVox~\cite{fang2022fast} approach introduced two versions of their model: The high quality base version (B), and a faster small version (S) that reduces the obtainable image quality in favor of computational speed.
As shown in the quantitative and qualitative comparisons, MoNeRF converges faster than TiNeuVox-S while achieving en-par quality with TiNeuVox-B (in a fraction of the training time), thus providing the best overall trade-off between training speed and final image quality. 
\cref{fig:convergence} provides detailed evidence of model convergence.  
MoNeRF's deformation and canonical components require significantly fewer training iterations than other methods using implicit scene representations, and yields the most detailed novel views in as few as $30$K update steps. 
%
\subsection{Real-World Scenes}
\begin{figure*}[ht]
    \centering
    \includegraphics[width=\textwidth,keepaspectratio]{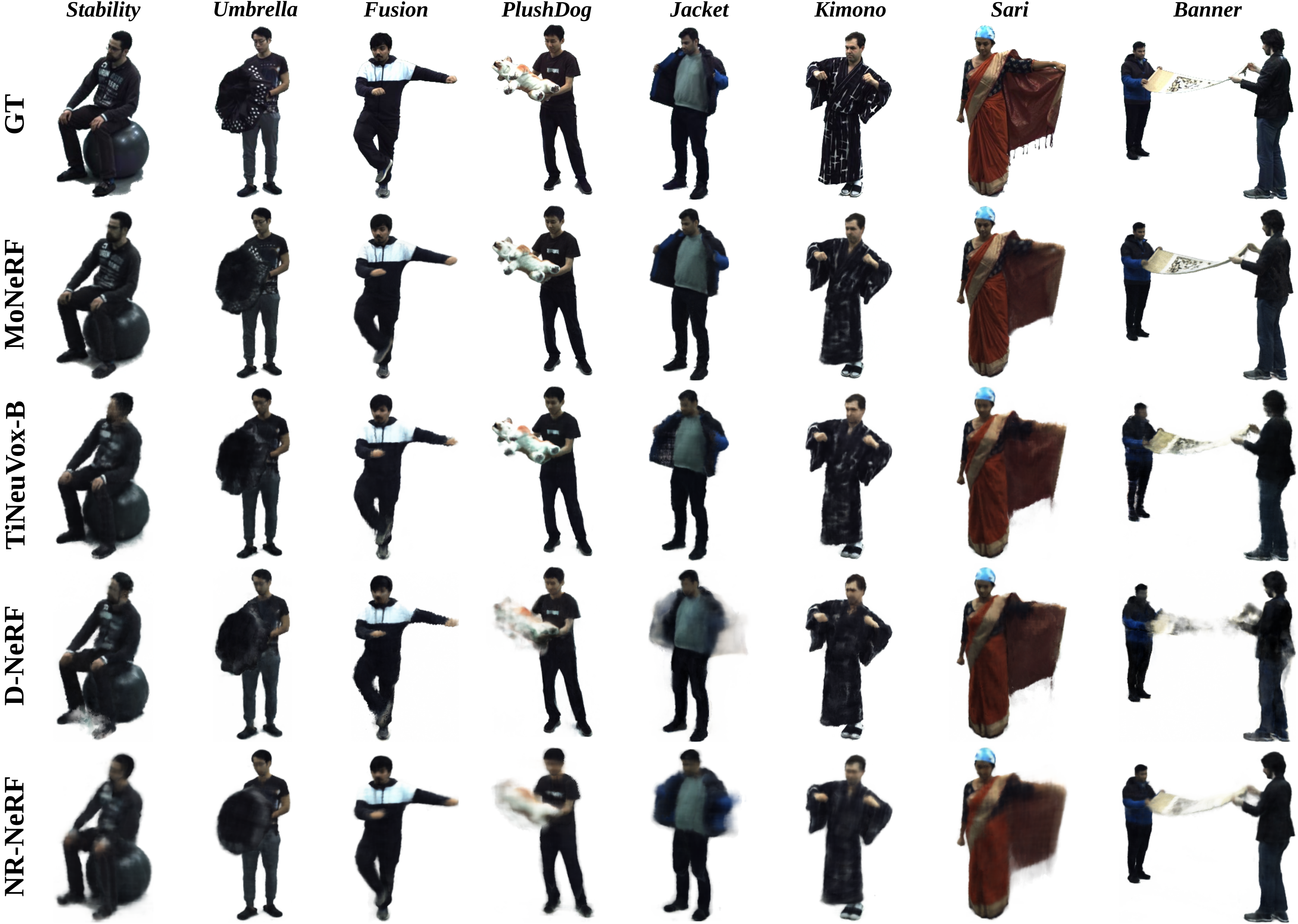}
    \caption{
        \textbf{Comparison on our MMVA dataset}. We compare our results (MoNeRF) against ground-truth (GT) and benchmark methods on a variety of real-world sequences comprising complex textures and motion.
        Note that MoNeRF produces sharper renderings 
        and finer details than other methods ({\eg} textures of deformable objects, such as garment wrinkles, the umbrella, and the plush toy).
        } 
    \label{fig:CompMMVA} 
\end{figure*} 
\begin{table}[th]
    \caption{
    \textbf{Quantitative comparison on our MMVA Dataset}. We report the averages according to PSNR/SSIM (higher is better) and LPIPS (lower is better) on the set of $12$ recorded scenes, as well as average training time and frames per second (FPS) during inference.} 
    \label{tab:AvgMetricsMMVA}
    \centering
    \footnotesize{
    \setlength\tabcolsep{5.5pt}
    \begin{tabular}{l|ccc|cc|}
        \multirow{2}{*}{\textbf{Method}} & \multicolumn{5}{c|}{\textbf{MMVA Dataset}}\\
        & \textbf{PSNR $\uparrow$} & \textbf{SSIM $\uparrow$} & \textbf{LPIPS $\downarrow$} & \textbf{Time} & \textbf{FPS}\\ 
        \hline
        InstantNGP~\cite{mueller2022instant} & 16.794 & 0.924 & 0.231 & 5 min & 75\\ 
        \hline
        D-NeRF~\cite{pumarola2021d} & 32.547 & 0.987 & 0.055 & $\sim$ h & 0.02\\ 
        NR-NeRF~\cite{tretschk2021non} & 31.029 & 0.986 & 0.020 & $\sim$ h & 0.06\\ 
        TiNeuVox-S~\cite{fang2022fast} & 34.411 & 0.991 & 0.014 & 9 min & 0.36\\ 
        TiNeuVox-B~\cite{fang2022fast} & 34.426 & 0.991 & 0.014 & 28 min & 0.19\\ 
        \hline
        MoNeRF (ours) & \textbf{35.079} & \textbf{0.993} & \textbf{0.011} & 7 min & 51
    \end{tabular}
    }
\end{table}

We examine the effectiveness of our approach on real-world data by performing quantitative and qualitative evaluations on our new MMVA dataset (introduced in \cref{sec:dataset}). 
In contrast to the low resolution synthetic data, we train all methods at full resolution of $1028\times752$ pixels. 
\cref{fig:CompMMVA} shows results for MoNeRF and related benchmark methods on a variety of sequences featuring challenging non-rigid deformations and texture patterns.
As evident from the qualitative evaluation -- despite unavoidable uncertainties arising from camera calibration, synchronization and foreground extraction -- monocularized data can effectively be applied for fast, high-fidelity non-rigid reconstruction in real-world scenarios.
Again, our method outperforms recent approaches both in image quality and training speed (see \cref{tab:AvgMetricsMMVA}).
Due to the efficient underlying hash encoding and acceleration structures, our method is the only one to achieve close to 60 FPS during novel-view rendering, making it applicable for real-time applications.
Similar to the synthetic data, MoNeRF reconstructs finer, more accurate object surfaces and texture details on a per-frame basis.
When comparing the quality of reconstructed scene motion (see our supplemental video), the static test camera view of the MMVA dataset uncovers significant temporal jittering for the implicit D-NeRF and NR-NeRF approaches, which was hidden by the spatially varying test camera views of the synthetic D-NeRF dataset.
In contrast, the explicit radiance field representations of TiNeuVox and MoNeRF provide better temporal regularization, resulting in smoother, temporally coherent video results.
Overall, we find that our MoNeRF -- without any adjustments or overhead -- 
translates well to real-world monocularized data, producing high-quality results in less than seven minutes while being robust to natural inaccuracies in multi-view camera calibration.

\subsection{Ablation Study}
\begin{table}[ht]
\begin{center}
    \caption{\textbf{Ablation experiment} on the effect of our deformation module factorization, the number of samples $q$ for updating the temporal occupancy grid, and the regularization losses. The highlighted rows show our final configuration.}
    \label{tab:ablation}
    \footnotesize{
    \setlength\tabcolsep{2pt}
    \begin{tabular}{c|c|c|c||c c c}
        \multirow{2}{1.4cm}{\centering \textbf{Temporal samples ($q$)}}& \multirow{2}{1cm}{\centering \textbf{Factori- zation}} & \multirow{2}{*}{ \centering\textbf{PSNR $\uparrow$}} & \multirow{2}{1.1cm}{\centering \textbf{Training Time $\downarrow$}} &
        \multicolumn{1}{c|}{\multirow{2}{.8cm}{\centering $\boldsymbol{L}_{\textrm{bg}}$ \cref{eq:lbg}}} & \multicolumn{1}{c|}{\multirow{2}{.8cm}{\centering $\boldsymbol{L}_{\textrm{def}}$ \cref{eq:ldef}}} & \multicolumn{1}{c|}{\multirow{2}{*}{\centering \textbf{PSNR $\uparrow$}}} \\ 
        & & & & \multicolumn{1}{c|}{} & \multicolumn{1}{c|}{} & \multicolumn{1}{c|}{}\\
         \hline
         1 & $\cmark$ & 31.88 & 4m 01s & \multicolumn{1}{c|}{$\xmark$} & \multicolumn{1}{c|}{$\xmark$} & \multicolumn{1}{c|}{31.35}  \\
         1 & $\xmark$ & 31.55 & 3m 57s & \multicolumn{1}{c|}{$\xmark$} & \multicolumn{1}{c|}{$\cmark$} & \multicolumn{1}{c|}{32.07}  \\
         \cellcolor{LightCyan}20 & \cellcolor{LightCyan}$\cmark$ & \cellcolor{LightCyan}32.16 & \cellcolor{LightCyan}6m 13s & \multicolumn{1}{c|}{$\cmark$} & \multicolumn{1}{c|}{$\xmark$} & \multicolumn{1}{c|}{31.52}  \\
         20 & $\xmark$ & 31.62 & 8m 18s & \multicolumn{1}{c|}{$\cellcolor{LightCyan}\cmark$} & \multicolumn{1}{c|}{\cellcolor{LightCyan}$\cmark$} & \multicolumn{1}{c|}{\cellcolor{LightCyan}32.16} \\
         30 & $\cmark$ & 32.23 & 7m 15s & & &\\
         30 & $\xmark$ & 31.85 & 10m 27s & & &\\
    \end{tabular}
    }
    \end{center}
\end{table}
We further study the impact of individual MoNeRF components. 
First, we investigate the significance of our deformation module factorization for the disentanglement of spatial and temporal information (\cref{sec:deformation_module}). 
To this end, we train an alternative version of MoNeRF that directly infers pointwise offset vectors from 4D spatiotemporal input samples in a single MLP with the same total amount of hidden layers.
Moreover, we experiment with the number of temporal samples $q$ used for approximating the temporal occupancy grid (\cref{subsec: acceleration}).
\cref{tab:ablation} shows training times and average PSNR on the D-NeRF dataset for different configurations.
We observe that our novel factorization-based deformation module achieves higher visual quality and enables faster occupancy grid updates (and thus overall training times) for larger $q$ values, which are needed to retain delicate scene appearance. 
Note that for $q>30$, most GPUs run out of memory for parallel execution, resulting in a significant drop in performance without notably improving the reconstruction quality. 
We also assess the influence of our regularization losses and observe that both losses contribute to the overall accuracy in terms of average PSNR. 
The best results are obtained using our full objective function.
\subsection{Additional Experiments}
We perform additional experiments on different types of data to further investigate the capabilities of our method and the applicability of multi-view monocularization in general.
\subsubsection{Comparison to Static Model}
\begin{figure}[t!]
    \centering
    \includegraphics[width=\columnwidth,keepaspectratio]{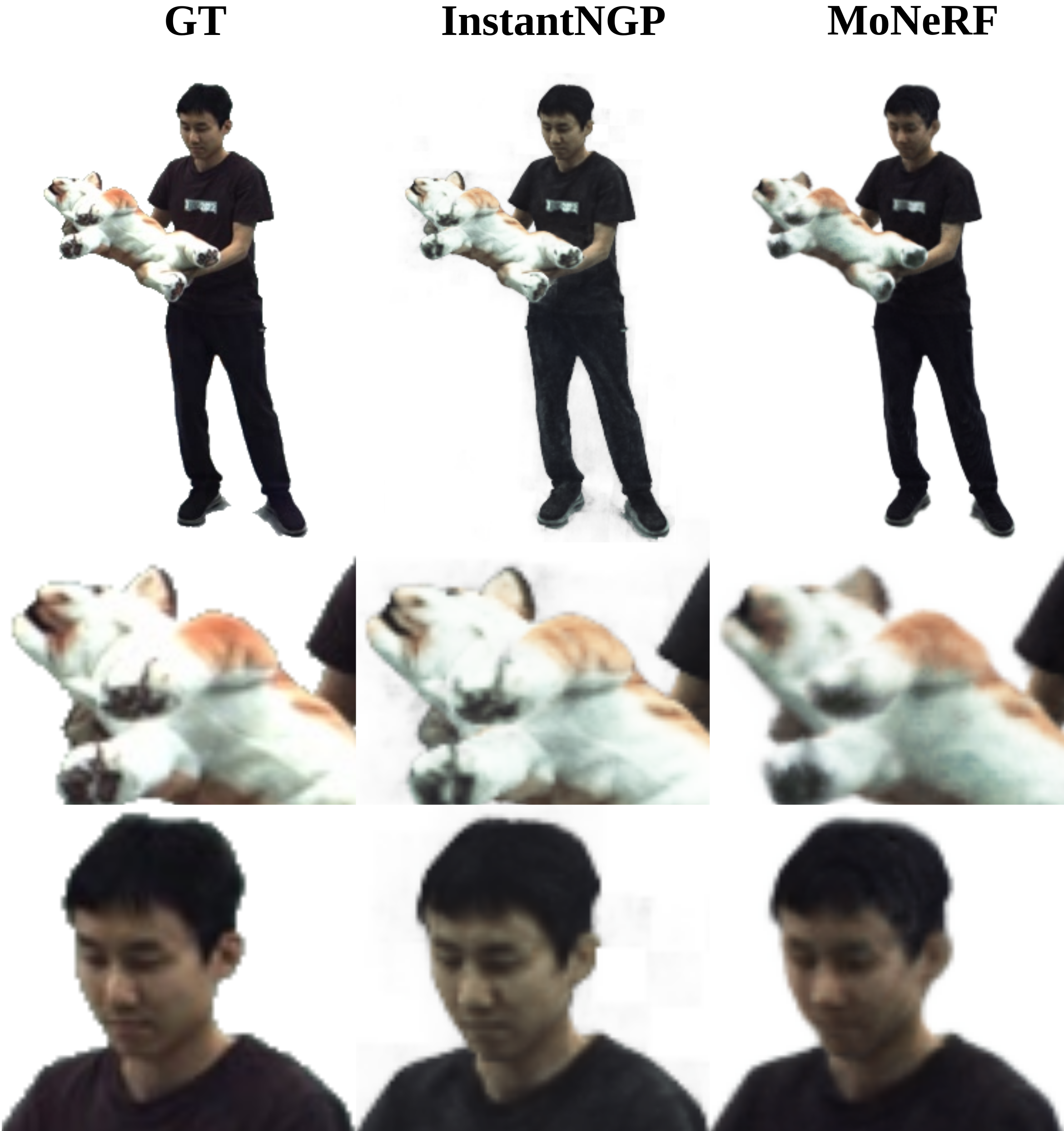} 
    \caption{\textbf{Comparison to single frame optimization:} We show a novel test view generated by InstantNGP~\cite{mueller2022instant} and our MoNeRF. While our method was trained on the monocularized dynamic sequence, InstantNGP was provided with $107$ multi-view images for the static frame. MoNeRF produces slightly blurry results in deforming areas of the scene, but maintains an overall high image quality.
    } 
    \label{fig:SUP-staticframe} 
\end{figure} 
To investigate the maximum capabilities of our model, we compare MoNeRF to the static neural radiance field implementation of InstantNGP~\cite{mueller2022instant}, which uses the same representation and parameters to model the scene.
To this end, we train InstantNGP on a single timestamp in the middle of one of our MMVA dataset sequences (PlushDog) using all $107$ multi-view cameras.
Our MoNeRF on the one hand was trained on a single camera of all $150$ timestamps, {\ie} has only seen one of the $107$ camera views of the static model, and is tasked to reconstruct the static object state by simultaneously inferring the deformation over the temporal sequence.
\cref{fig:SUP-staticframe} shows renderings of the test view unseen by both methods during training.
This experiment explores the upper bound of our model capabilities, as our dynamic reconstruction from monocularized data naturally cannot improve over full static per-frame multi-view reconstruction.
We observe that our MoNeRF produces more blur in dynamic image regions, like the plush dog and the actor's face, as it has to jointly infer the scene motion and appearance from a single image at a time.
InstantNGP, on the other hand, when provided with multi-view data, can reconstruct slightly finer details and less blur on the object's surface.
However, while introducing marginal blur to the final renderings, our full dynamic MoNeRF model is trained in under seven minutes on less than $1\%$ of the training data, while multi-view InstantNGP training takes $\sim$$5$ minutes per frame, resulting in $12.5$ hours of training and a significant increase in training data and memory usage to reconstruct the entire temporal sequence.
\subsubsection{Monocularized Stereo}
\begin{figure}[ht]
    \centering
    \includegraphics[width=\columnwidth,keepaspectratio]{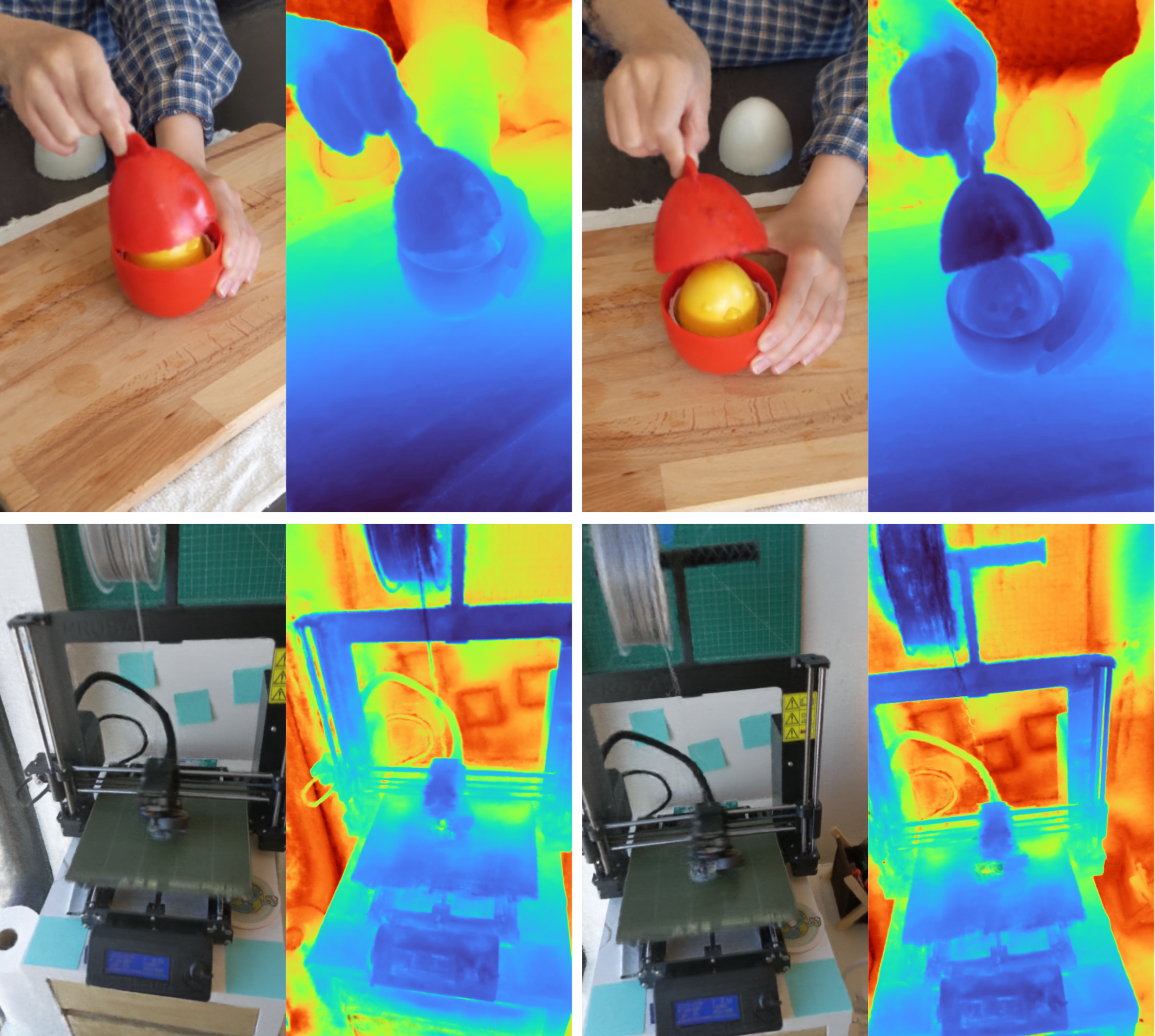}
    \caption{
        \textbf{Qualitative results on monocularized stereo recordings}. We show results on the HyperNeRF~\cite{park2021hypernerf} sequences captured by a stereo setup and trained in a monocularized setup. For two exemplary scenes (chicken and 3D printer), we show two test frames at different viewpoint and timestamps, and the according depth map.
}    
    \label{fig:CompHyperNeRF} 
\end{figure} 
\begin{table*}[ht]
    \caption{\textbf{Quantitative comparison on the HyperNeRF dataset}. We train our model on the four dynamic \textit{vrig} scenes. Values for other methods are copied from the corresponding papers.} 
    \label{tab:compdhypernerf}
    \centering
    \footnotesize{
    \setlength\tabcolsep{3pt}
    \begin{tabular}{l|cccccccc|cc|c}
        \multirow{2}{*}{\textbf{Method}} & \multicolumn{2}{c}{\textbf{Broom}} & \multicolumn{2}{c}{\textbf{3D Printer}} & \multicolumn{2}{c}{\textbf{Chicken}} & \multicolumn{2}{c|}{\textbf{Peel Banana}} & \multicolumn{2}{c|}{\textbf{Mean}} & \textbf{Time}\\
        & \textbf{PSNR} $\uparrow$ & \textbf{\textbf{MS-SSIM} $\uparrow$} & \textbf{PSNR} $\uparrow$ & \textbf{\textbf{MS-SSIM} $\uparrow$} & \textbf{PSNR} $\uparrow$ & \textbf{\textbf{MS-SSIM} $\uparrow$} & \textbf{PSNR} $\uparrow$ & \textbf{\textbf{MS-SSIM} $\uparrow$} & \textbf{PSNR} $\uparrow$ & \textbf{\textbf{MS-SSIM} $\uparrow$} & \\ 
        \hline
        NSFF~\cite{li2021neural} & \textbf{26.1} & \textbf{0.871} & \textbf{27.7} & \textbf{0.947} & 26.9 & 0.944 & \textbf{24.6} & \textbf{0.902} & \textbf{26.3} & \textbf{0.916} & $\sim$ h \\
        Nerfies~\cite{park2021nerfies} & 19.2 & 0.567 & 20.6 & 0.830 & 26.7 & 0.943 & 22.4 & 0.872 & 22.2 & 0.803 & $\sim$ h \\
        HyperNeRF~\cite{park2021hypernerf} & 19.3 & 0.591 & 20.0 & 0.821 & 26.9 & 0.948 & 23.3 & 0.896 & 22.4 & 0.814 & $\sim$ h \\
        \hline
        TiNeuVox-S~\cite{fang2022fast} & 21.9 & 0.707 & 22.7 & 0.836 & 27.0 & 0.929 & 22.1 & 0.780 & 23.4 & 0.813 & $10$ min \\
        TiNeuVox-B~\cite{fang2022fast} & 21.5 & 0.686 & 22.8 & 0.841 & 28.3 & 0.947 & 24.4 & 0.873 & 24.3 & 0.837 & $30$ min \\
        \hline
        MoNeRF (ours) & 21.6 & 0.665 & 22.3 & 0.835 & \textbf{29.9} & \textbf{0.966} & 24.4 & 0.879 & 24.6 & 0.836 & $15$ min \\
    \end{tabular}
    }
\end{table*}
Similar to the synthetic D-NeRF dataset~\cite{pumarola2021d}, the HyperNeRF dataset by Park \etal~\cite{park2021hypernerf} features another interesting implementation of data monocularization.
It features multiple real-world scenes recorded as a continuous trajectory using two smartphones installed on a stereo rig.
For training, consecutive images are sampled from the left and right camera in an alternating fashion, resulting in a monocular training sequence with effective multi-view signal.
However, while our method was designed to reconstruct entire objects from all sides (without background) from large 360{\textdegree} inward-facing capturing setups, the HyperNeRF scenes were recorded using a single, mostly forward-facing camera path.
Despite this discrepancy in application, we evaluate our method on the four \textit{vrig} recordings.
Qualitative and quantitative results are provided in \cref{fig:CompHyperNeRF} and \cref{tab:compdhypernerf}, respectively.
As shown, we achieve promising preliminary results without any changes to our pipeline, only adding the efficient distortion loss presented by MipNeRF360~\cite{barron2022mipnerf360} to regularize the depth due to the forward facing setup.
While our method does not reach state-of-the-art performance, it performs en-par with TiNeuVox~\cite{fang2022fast} in terms of average PSNR.
More detailed qualitative analysis yields that our method produces outstanding results on scenes with regular geometry (e.g. the chicken), where it preserves finer details, but sometimes fails to reconstruct fine geometric details or produces blur for more complex geometry like the 3D printer.
Thus, we conclude that the deformation modeling of TiNeuVox is currently better suited for forward-facing few camera setups, even though fine textural details are lost.
\subsubsection{Sparse Multi-View Data}
\begin{figure}[t!]
    \centering
    \includegraphics[width=\columnwidth,keepaspectratio]{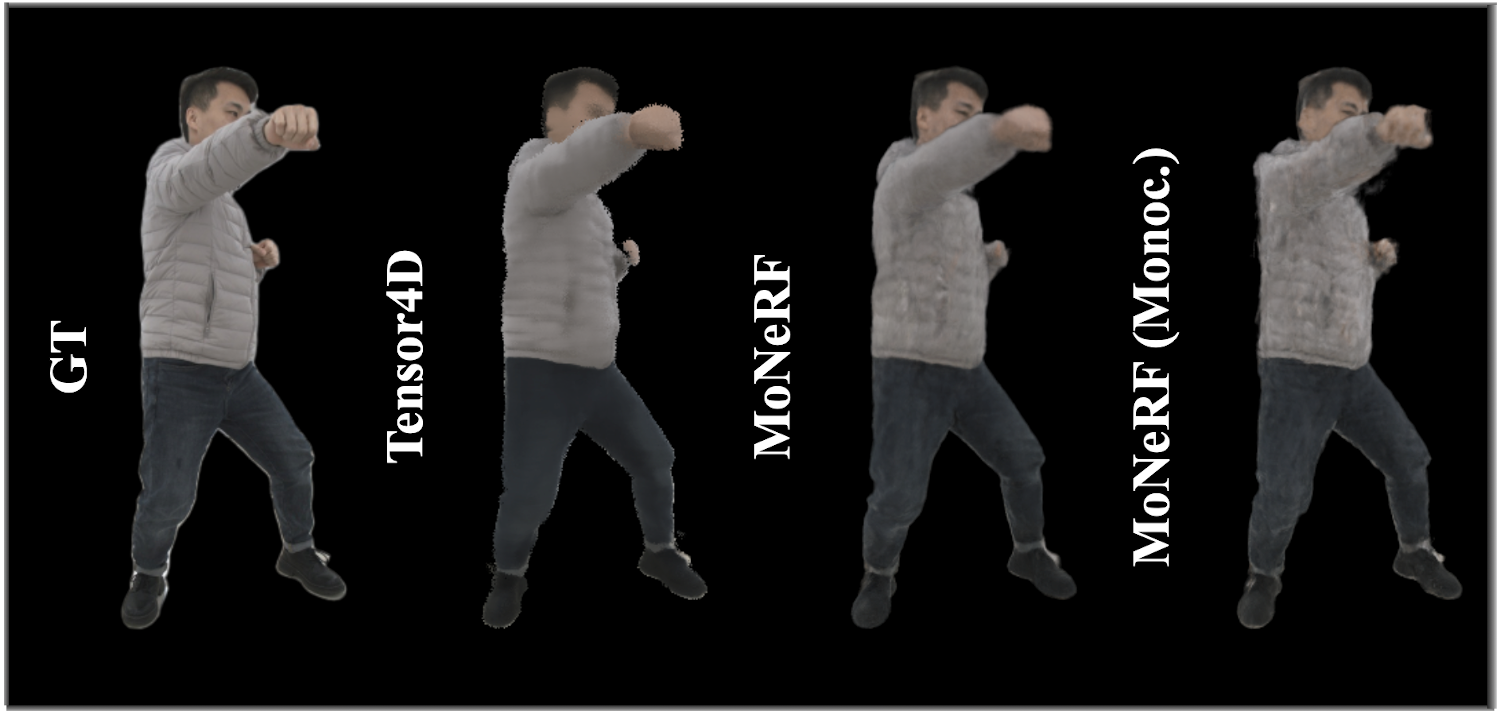}
    \footnotesize{
    \setlength\tabcolsep{3pt}
    \begin{tabular}{l|ccc|cc}
    \multirow{2}{*}{\textbf{Method}} & \multicolumn{3}{c|}{\textbf{Image Metrics}} & \multicolumn{2}{c}{\textbf{Training}}\\
    & \textbf{PSNR $\uparrow$} & \textbf{SSIM $\uparrow$} & \textbf{LPIPS $\downarrow$} & \textbf{Time} & \textbf{Images}\\ 
    \hline
    Tensor4D~\cite{shao2023tensor4d} & 29.36 & 0.95 & 0.10 & 15 h & 330 \\ 
    MoNeRF & 29.89 & 0.95 & 0.07 & 8 min & 330 \\
    MoNeRF (Monoc.) & 27.84 & 0.93 & 0.08 & 8 min & 30  \\
    \end{tabular}
    }
    \caption{
    \textbf{Comparison on sparse multi-view data}. We train our method on the sparse multi-view boxing sequence provided by Tensor4D~\cite{shao2023tensor4d}. Similar to the MMVA sequences, MoNeRF produces detailed novel views after only 8 minutes of training. Here, in contrast to large-scale multi-view rigs, data monocularization offers less effective  benefits, but introduces a higher error to the final reconstruction due to the dataset's low temporal resolution.
    }\label{fig:CompTensor4d} 
\end{figure}

We further test our MoNeRF on sparse multi-view data, {\ie} full 360{\textdegree} inward-facing recordings with few spatial viewpoints and a low temporal resolution.
In contrast to our MMVA recordings, this type of data naturally only provides a limited amount of training views with small memory footprint, and can thus efficiently be processed without the need for monocularization.
To this end, we train our model on the boxing sequence taken from the data samples provided by the factorization-based Tensor4D~\cite{shao2023tensor4d} approach.
This sequence features only $12$ synchronized cameras for the full hemisphere, and a total of $30$ timestamps with a low temporal resolution ({\ie} larger deformations over time).
We leave out one camera for testing, resulting in a total of $330$ training views.
As shown in \cref{fig:CompTensor4d}, MoNeRF can leverage the multi-view information to render detailed novel views after only $8$ minutes of training.
Even though data monocularization does not provide any benefits on compact-sized sparse multi-view data, we test the limits of this procedure by training on a single, randomly sampled view per timestamp, resulting in only $30$ input training views.
We find that, in the case of low temporal resolution recordings with few cameras, monocularization can lead to a more noticeable drop in visual quality and motion fidelity, indicating that this technique is best suited for large-scale multi-view recordings.
\subsubsection{Forward-Facing Multi-View Scenes}
\begin{figure*}[ht]
    \centering
    \includegraphics[width=\textwidth,keepaspectratio]{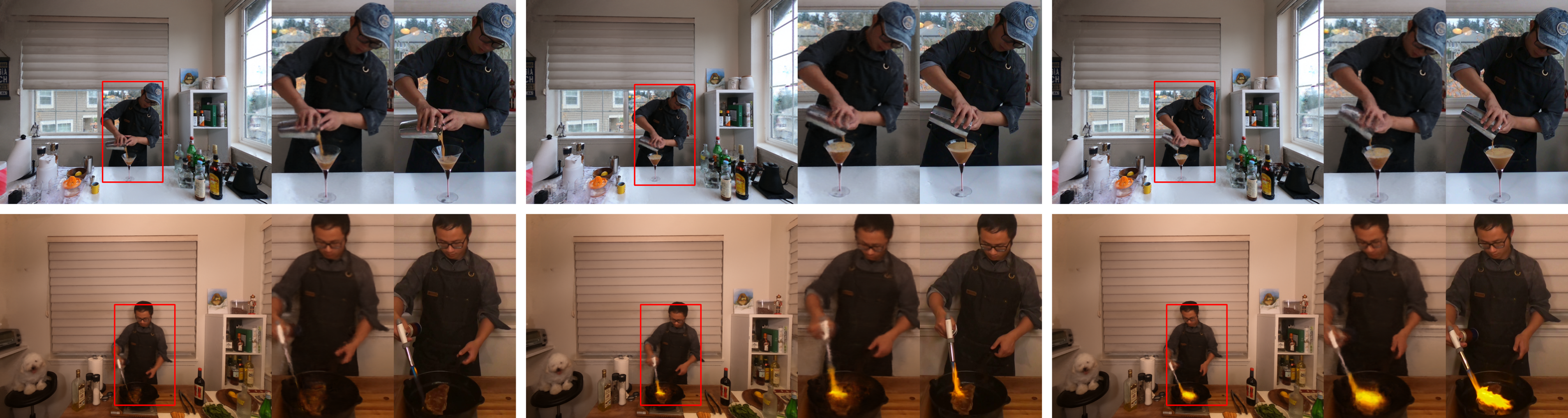}
    \caption{
        \textbf{Qualitative results on forward-facing scenes}. We show results of our MoNeRF trained on two monocularized sequences of the DyNeRF~\cite{li2022neural} dataset. For every frame, we provide closeup comparison of our result (left) to GT (right) on the dynamic foreground. 
}    
    \label{fig:CompDyNeRF} 
\end{figure*} 
\begin{table*}[ht]
    \caption{
    \textbf{Quantitative comparison on the DyNeRF dataset}. We train our MoNeRF on the six dynamic forward-facing DyNeRF~\cite{li2022neural} sequences. For MoNeRF (ours) and MixVoxels-L~\cite{wang2022mixed}, we also report values for the model trained on a monocularized version of the dataset. Results and training times of other approaches are copied from the corresponding papers.} 
    \label{tab:compdynerf}
    \centering
    \footnotesize{
    \setlength\tabcolsep{5pt}
    \begin{tabular}{l|ccccccc|cc}
        \multirow{2}{*}{\textbf{Method}} & \multicolumn{7}{c|}{\textbf{PSNR} $\uparrow$} & \multicolumn{2}{c}{\textbf{Training}}\\
        & \textbf{Coffee Martini} & \textbf{Cook Spinach} & \textbf{Cut Beef} & \textbf{Flame Salmon}\textsuperscript{1} & \textbf{Flame Steak} & \textbf{Sear Steak}& \textbf{Mean} & \textbf{Time} & \textbf{Images}\\ 
        \hline
        $K$-Planes-hybrid~\cite{kplanes_2023} & 28.74 & 32.19 & 31.93 & 28.71 & 31.80 & 31.89 & 30.88 & 4 h & $\sim$6000 \\
        DyNeRF~\cite{li2022neural} & -- & -- & -- & 29.58 & -- & -- & -- & 1344 h & $\sim$6000 \\
        MixVoxels-L~\cite{wang2022mixed} & 29.36 & 31.61 & 31.30 & 29.92 & 31.21 & 31.43 & 30.80 & 1.3 h & $\sim$6000 \\
        MixVoxels-L\textsuperscript{2}~\cite{wang2022mixed} & 28.46 & 28.39 & 28.51 & 28.33 & 28.61 & 30.32 & 28.77 & 1.3 h & 300 \\
        MoNeRF & 28.08 & 31.10 & 32.12 & 27.41 & 32.28 & 32.75 & 30.62 & 20 min\textsuperscript{3} & $\sim$6000 \\
        MoNeRF\textsuperscript{2} & 28.28 & 30.63 & 31.48 & 27.19 & 31.40 & 32.50 & 30.25 & 15 min\textsuperscript{3} & 300 \\
        \hline
         & \multicolumn{7}{c|}{\textbf{SSIM} $\uparrow$} & & \\
         & \textbf{Coffee Martini} & \textbf{Cook Spinach} & \textbf{Cut Beef} & \textbf{Flame Salmon}\textsuperscript{1} & \textbf{Flame Steak} & \textbf{Sear Steak}& \textbf{Mean} & & \\ 
        \cline{1-8}
        $K$-Planes-hybrid~\cite{kplanes_2023} & 0.953 & 0.966 & 0.966 & 0.953 & 0.970 & 0.974 & 0.964 & & \\
        DyNeRF~\cite{li2022neural} & -- & -- & -- & 0.961 & -- & -- & -- &  & \\
        MixVoxels-L~\cite{wang2022mixed} & 0.946 & 0.965 & 0.965 & 0.945 & 0.970 & 0.971 & 0.960 &  & \\
        MixVoxels-L\textsuperscript{2}~\cite{wang2022mixed} & 0.939 & 0.952 & 0.945 & 0.939 & 0.961 & 0.966 & 0.950 &  & \\
        MoNeRF & 0.879 & 0.927 & 0.933 & 0.869 & 0.940 & 0.943 & 0.915 &  & \\
        MoNeRF\textsuperscript{2} & 0.882 & 0.923 & 0.928 & 0.864 & 0.935 & 0.941 & 0.912 &  & \\
        \cline{1-8}
        \multicolumn{10}{c}{} \\
        \multicolumn{5}{l}{\textsuperscript{1}Using only the first 10 seconds of the full video} & \multicolumn{5}{l}{\textsuperscript{2}Trained on monocularized dataset version} \\ 
        \multicolumn{5}{l}{\textsuperscript{3}Times include dataloading} & \multicolumn{5}{l}{``--'' denotes unreported values.} \\
    \end{tabular}
    }
\end{table*}
While our method is designed for fast full 360{\textdegree} dynamic object rendering, another branch of methods tackles the reconstruction from real-world forward-facing multi-view recordings.
One of the most prominent datasets for this task was introduced by DyNeRF~\cite{li2022neural}, featuring $6$ scenes with a length of $10$ seconds, recorded by up to $20$ cameras at $2$K resolution with $30$Hz.
As forward-facing scenes bring their own unique challenges, we test the limits of our approach by training our MoNeRF on the DyNeRF dataset.
Similar to previous work, we downsample the videos to half resolution, resulting in a total of $\sim6000$ training views at $1$K resolution.
To improve the model performance for forward-facing scenes, we enhance our model with the efficient distortion loss introduced by Mip-NeRF~360~\cite{barron2022mipnerf360} to regularize the estimated per-pixel depth, and apply per-camera latent codes as suggested by InstantNGP~\cite{mueller2022instant} to compensate for the varying camera color calibrations.
To further test the practicability and influence of data monocularization in such scenarios, we also train our MoNeRF, and the recent state-of-the-art multi-view method MixVoxels~\cite{wang2022mixed} on a monocularized version ({\ie} $300$ images) of the dataset.
We show qualitative novel view results in \cref{fig:CompDyNeRF}, and provide quantitative comparisons in \cref{tab:compdynerf}.
As evident from the results, the monocularized version of our MoNeRF is efficient in training while maintaining a high level of quality, but introduces blur in the dynamic foreground. 
This blur can only partially be corrected by adding all images from the forward-facing multi-view setup, as the domain mismatch continues to exist.
On the other hand, we observe a drop in image quality when training MixVoxels on monocularized data.
This reduction of quality stems from a severe loss in temporal stability, resulting in jitter in the dynamic foreground, while the static background remains mostly unimpaired.
We conclude that our MoNeRF currently cannot fully leveraging multi-view signals in forward-facing recordings, while the multi-view method MixVoxels can not reconstruct coherent motion from monocularized sequences, as the approaches were not developed or optimized for the respective scenarios.
In the future, dedicated optimizations of our model and acceleration structures for parallel camera views can help our method to further increase the performance in this scenario.
\subsubsection{Continuous Camera Trajectories}
\begin{figure}[b!]
    \centering
    \includegraphics[width=1.0\linewidth, keepaspectratio]{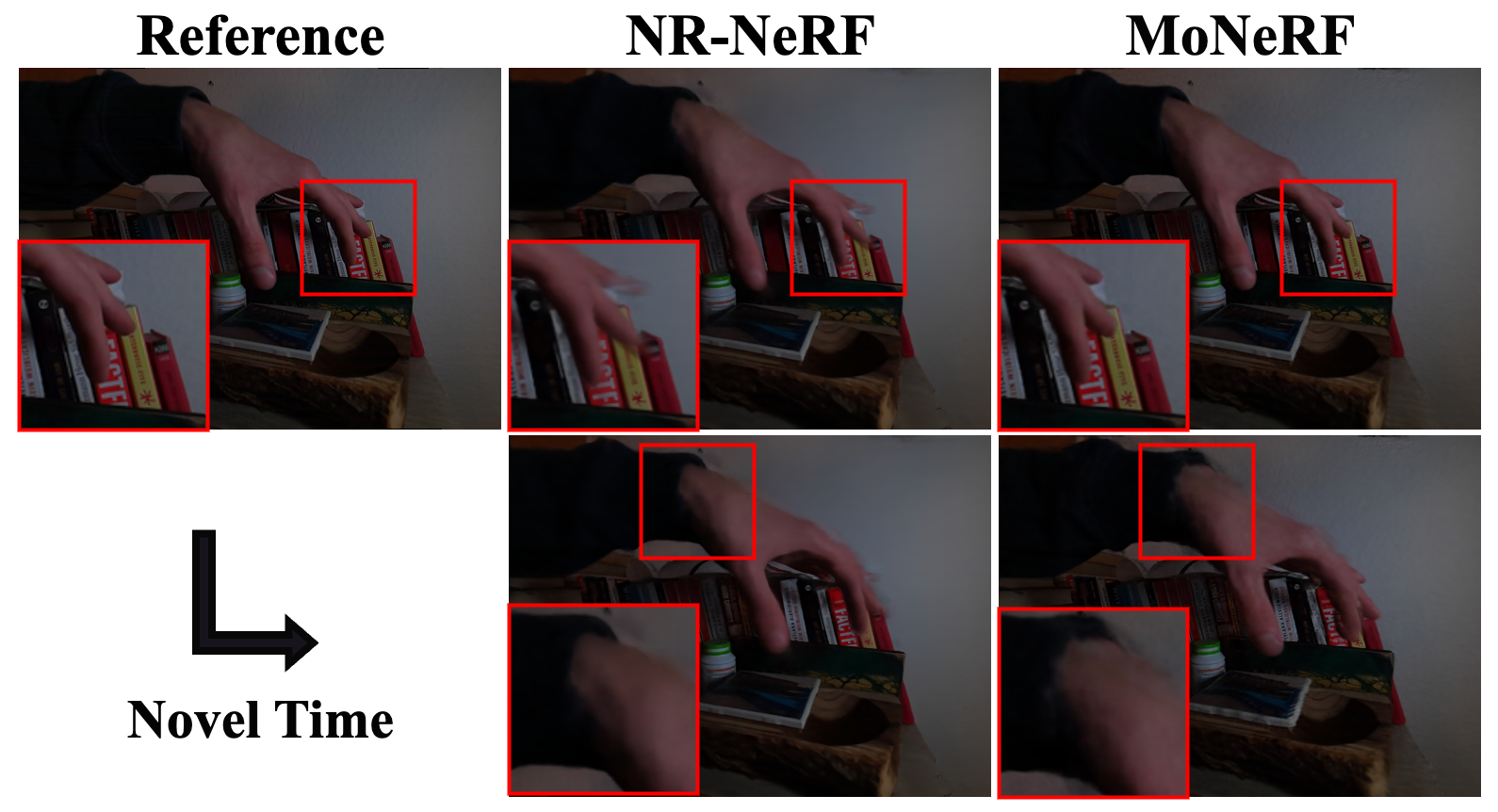}
    \caption{
        \textbf{Comparison on continuous camera trajectory.} 
        We train our model on a continuous sequence provided by NR-NeRF~\cite{tretschk2021non}, and synthesize novel timestamps from a static training view. 
        (Top): The training view. (Bottom): A novel timestamp that was not observed from this camera position. 
    } 
    \label{fig:comp_nrnerf} 
\end{figure} 
One of the most difficult applications of dynamic radiance fields is the reconstruction and interpolation of a single video, {\ie} a continuous monocular camera trajectory.
While most other approaches are designed for and tested on monocularized sequences with an effective multi-view signal, NR-NeRF~\cite{tretschk2021non} shows promising results on continuous camera footage.
To further test the capabilities of our model (design for monocularized inward-facing setups), we train MoNeRF on a publicly available training sequence by Tretschk~\etal~\cite{tretschk2021non}.
We find that, while MoNeRF preserves finer geometric details, the generalization of motion to far off camera views is currently limited, as shown in \cref{fig:comp_nrnerf}.
We assume that these effects are caused by the explicit nature of the underlying hash-grid representation, which does not offer the natural continuous regularization of large MLPs.
Similar to forward-facing scenes, future investigation of dedicated regularizers can help to improve the performance of MoNeRF in such scenarios.
\section{Conclusion}\label{sec:conclusion} 
We introduced MoNeRF, an unprecedentedly fast and accurate method for novel view synthesis of arbitrary non-rigidly deforming scenes in a monocularized 360{\textdegree} inward-facing setting. 
The win-win combination of fast optimization and accurate real-time novel-view synthesis is enabled by the new deformation module with separate spatial and temporal components, in combination with the fast hash-encoded canonical scene representation. 
Our experiments show that MoNeRF can reconstruct qualitatively appealing radiance and (temporally coherent) deformation fields for challenging scenes with fine appearance details in a matter of minutes, and render novel views at real-time framerates, \textit{i.e.,} substantially faster than previous approaches. 
Moreover, we observe in the tests with our new dataset that the advantages of the proposed approach and monocularized setting can directly be transferred to real-world recordings, and even yield promising results in other setups such as forward-facing camera rigs or fully monocular video.
In the future, explicit spatial and temporal regularizers can be explored to further enhance the performance in those scenarios.
Likewise, a keyframe-based model in the style of Nerfplayer~\cite{song2023nerfplayer} can be integrated to extend the representable scene length, which is currently limited to about $10$ seconds.
Furthermore, our new dataset enables the investigation of more sophisticated importance sampling techniques~\cite{pan2022activenerf} on ray- or image-level that can be applied to optimize the setting for practical purposes.
Thus, we believe our work opens many opportunities for applications and further research.

\section*{Acknowledgments}
This work was partially funded by the DFG (MA2555/15-1 ``Immersive Digital Reality'') and the ERC Consolidator Grant 4DRepLy (770784).
We thank Florian Hahlbohm, Timon Scholz, Basavaraj R. Sunagad, and Kamuni Pranay Raj for helping with data recording and comparisons.

\putbib
\end{bibunit}
%
\clearpage
\setcounter{section}{0}
\renewcommand{\thesection}{S\Roman{section}}
\setcounter{figure}{0}
\renewcommand{\figurename}{Fig.}
\renewcommand{\thefigure}{S\arabic{figure}}
\setcounter{table}{0}
\renewcommand{\thetable}{S\Roman{table}}
%
\title{Fast Non-Rigid Radiance Fields\\ from Monocularized Data\\ {-- Supplementary Material --}}

\author{
Moritz Kappel$^1$\hspace{1.8em}
Vladislav Golyanik$^2$\hspace{1.8em}
Susana Castillo$^1$\hspace{1.8em}
Christian Theobalt$^2$\hspace{1.8em}
Marcus Magnor$^1$\vspace{0.8em}\\
{\parbox{\textwidth}{\centering \small $^1$ Computer Graphics Lab, TU Braunschweig, Germany
    \hspace{7pt}{\tt\small \{lastName\}@cg.cs.tu-bs.de}\\
    \small $^2$ Max Planck Institute for Informatics, Saarland Informatics Campus, Germany
    \hspace{7pt}{\tt\small \{lastName\}@mpi-inf.mpg.de}\vspace{-1cm}
      }
    }
}

\markboth{\tiny{This work has been submitted to the IEEE for possible publication. Copyright may be transferred without notice, after which this version may no longer be accessible.}}%
{Kappel \MakeLowercase{\textit{et al.}}: Fast Non-Rigid Radiance Fields from Monocularized Data}

\maketitle 
\begin{bibunit}
\textbf{This supplementary document provides more details on our new \textit{MMVA} dataset recording setup (Sec.~\ref{sec:MMVA}), along with per scene comparisons between the proposed \textit{MoNeRF} and the benchmark methods examined in the main paper (Sec.~\ref{sec:additional_results}).}

\section{Recording Setup}\label{sec:MMVA}
\begin{figure}[htb]
\centering
    \includegraphics[width=\columnwidth,keepaspectratio]{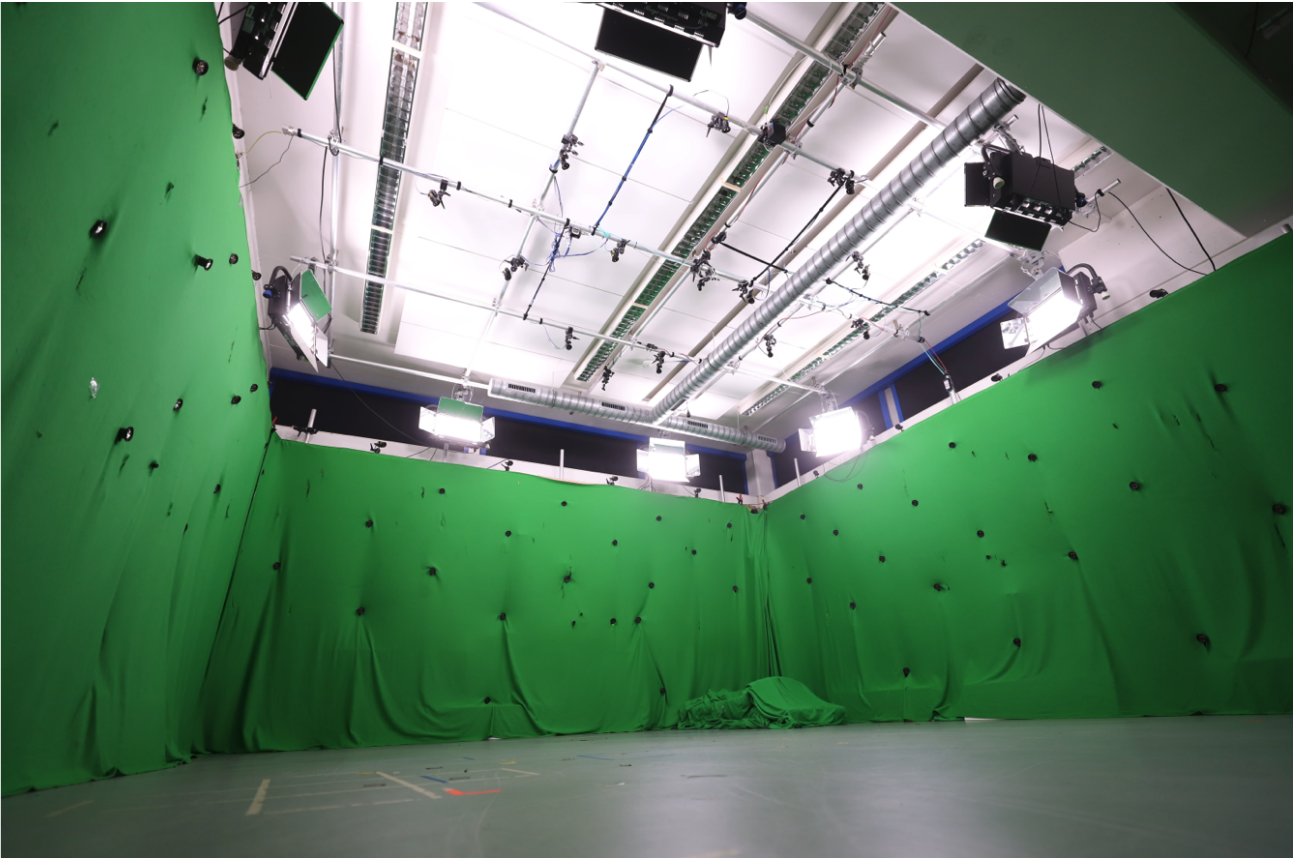}
    \captionof{figure}{ 
    \textbf{Multi-view recording setup for the MMVA dataset.}}
    \label{fig:SUP-RecordSetup} 
\end{figure}

When creating our MMVA dataset, we closely follow the D-NeRF dataset specifications to maintain comparability and compatibility with recent methods. 
As continuous sampling is impossible for real-world setups, and we want to avoid motion blur from fast-moving cameras, we record our dataset using a large-scale synchronized multi-view setup; see~\cref{fig:SUP-RecordSetup}.
We then monocularize the multi-view data by selecting a single camera image per timeframe for training, {\ie} a discrete version of the D-NeRF hemisphere sampling.
Using this setup, we record a total of twelve sequences showing human actors and general object interaction.
Like the D-NeRF sequences, our recordings contain single short motions ($100-250$ frames, $2-5$ seconds). 
A visualization of all our sequences is available in our supplemental video.
We calibrate the cameras using COLMAP~\cite{schoenberger2016sfm, schoenberger2016mvs} and extract foreground masks via background subtraction and human body segmentation.
We then re-scale the camera extrinsics such that the actor is located at a zero-centered cube of side $0.5$. 
The final data format is designed to be compatible with existing NeRF implementations and only adds per-camera focal lengths and principal points.
Before our recordings, we gathered consent forms from all actors, enabling us to release the full dataset for research purposes.

\section{Detailed Comparisons}\label{sec:additional_results}
\begin{table*}[ht]
    \caption{\textbf{Quantitative per-scene comparison on the D-NeRF~\cite{pumarola2021d} dataset}. We report PSNR/SSIM (higher is better) and LPIPS (lower is better) for all eight dynamic scenes of the D-NeRF dataset. The best results are highlighted in bold. 
    MoNeRF, InstantNGP, and NR-NeRF training times were measured on a single NVidia RTX 3090 GPU. Values for NeRF, DirectVoxGO, and Plenoxels are taken from Fang~\etal~\cite{fang2022fast}, and values for all other methods are taken from the respective original papers.
    }\label{tab:MetricsDNeRF}
    \centering
    \small{
    \setlength\tabcolsep{2pt}
    \begin{tabular}{lcccccccccccccccc}
        \hline
        &  & \multicolumn{3}{c}{\textit{Hell Warrior}} &  & \multicolumn{3}{c}{\textit{Mutant}} &  & \multicolumn{3}{c}{\textit{Hook}} &  & \multicolumn{3}{c}{\textit{Bouncing Balls}} \\
        \textbf{Method} &  & \textbf{PSNR $\uparrow$} & \textbf{SSIM $\uparrow$} & \textbf{LPIPS $\downarrow$} &  & \textbf{PSNR $\uparrow$} & \textbf{SSIM $\uparrow$} & \textbf{LPIPS $\downarrow$} &  & \textbf{PSNR $\uparrow$} & \textbf{SSIM $\uparrow$} & \textbf{LPIPS $\downarrow$} &  & \textbf{PSNR $\uparrow$} & \textbf{SSIM $\uparrow$} & \textbf{LPIPS $\downarrow$} \\ \cline{1-1} \cline{3-5} \cline{7-9} \cline{11-13} \cline{15-17} 
        NeRF~\cite{mildenhall2020nerf} &  & 13.52 & 0.81 & 0.25 && 20.31 & 0.91 & 0.09 && 16.65 & 0.84 & 0.19 && 20.26 & 0.91 & 0.20  \\ 
        DirectVoxGO~\cite{sun2022direct} &  & 13.32 & 0.75 & 0.25 && 19.45 & 0.89 & 0.12 && 16.16 & 0.80 & 0.21 && 20.20 & 0.87 & 0.22\\
        Plenoxels~\cite{yu2021plenoxels} &  &  15.19 & 0.78 & 0.27 && 21.44 & 0.91 & 0.09 && 17.90 & 0.81 & 0.21 && 21.30 & 0.89 & 0.18\\
        InstantNGP~\cite{mueller2022instant} && 15.28 & 0.84 & 0.26 && 20.59 & 0.91 & 0.11 && 15.92 & 0.82 & 0.23 && 19.12 & 0.89 & 0.19\\
        \cline{1-1} \cline{3-5} \cline{7-9} \cline{11-13} \cline{15-17} 
        T-NeRF~\cite{pumarola2021d} &&  23.19 & 0.93 & 0.08 && 30.56 & 0.96 & 0.04 && 27.21 & 0.94 & 0.06 && 32.01 & 0.97 & 0.04\\
        D-NeRF~\cite{pumarola2021d} &&  25.02 & 0.95 & \textbf{0.06} && 31.29 & 0.97 & 0.02 && 29.25 & 0.96 & 0.11 && 32.80 & 0.98 & \textbf{0.03}\\ 
        NR-NeRF~\cite{tretschk2021non} && 23.74 & 0.94 & 0.07 && 30.77 & 0.97 & 0.03 && 26.49 & 0.94 & 0.07 && 24.72 & 0.95 & 0.15\\
        TiNeuVox-S~\cite{fang2022fast}&& 27.00 & 0.95 & 0.09 && 31.09 & 0.96 & 0.05 && 29.30 & 0.95 & 0.07 && 39.05 & \textbf{0.99} & 0.06  \\
        TiNeuVox-B~\cite{fang2022fast}&&  \textbf{28.17} & \textbf{0.97} & 0.07 && 33.61 & 0.98 & 0.03 && \textbf{31.45} & 0.97 & 0.05 && 40.73 & \textbf{0.99} & 0.04\\
        NDVG~\cite{guo2022neural} && 26.49 & 0.96 & 0.07 && 34.41 & 0.98 & 0.03 && 30.00 & 0.96 & 0.05 && 37.52 & \textbf{0.99} & 0.08\\
        $K$-Planes-hybrid~\cite{kplanes_2023} && 25.70 & 0.95 & -- && 33.79 & 0.98 & -- && 28.50 & 0.95 & -- && \textbf{41.22} & \textbf{0.99} & --\\
        \textbf{MoNeRF} && 26.53 & 0.96 & \textbf{0.06} && \textbf{35.51} & \textbf{0.99} & \textbf{0.01} && 31.12 & \textbf{0.98} & \textbf{0.03} && 39.45 & \textbf{0.99} & 0.04\\  
        \hline
        &  & \multicolumn{3}{c}{} &  & \multicolumn{3}{c}{} &  & \multicolumn{3}{c}{} &  & \multicolumn{3}{c}{} \\
        &  & \multicolumn{3}{c}{\textit{Lego}} &  & \multicolumn{3}{c}{\textit{T-Rex}} &  & \multicolumn{3}{c}{\textit{Stand Up}} &  & \multicolumn{3}{c}{\textit{Jumping Jacks}} \\
        \textbf{Method} &  & \textbf{PSNR $\uparrow$} & \textbf{SSIM $\uparrow$} & \textbf{LPIPS $\downarrow$} &  & \textbf{PSNR $\uparrow$} & \textbf{SSIM $\uparrow$} & \textbf{LPIPS $\downarrow$} &  & \textbf{PSNR $\uparrow$} & \textbf{SSIM $\uparrow$} & \textbf{LPIPS $\downarrow$} &  & \textbf{PSNR $\uparrow$} & \textbf{SSIM $\uparrow$} & \textbf{LPIPS $\downarrow$} \\ \cline{1-1} \cline{3-5} \cline{7-9} \cline{11-13} \cline{15-17} 
        NeRF~\cite{mildenhall2020nerf} &  & 20.30 & 0.79 & 0.23 && 24.49 & 0.93 & 0.13 && 18.19 & 0.89 & 0.14 && 18.28 & 0.88 & 0.23\\
        DirectVoxGO~\cite{sun2022direct} &  & 21.13 & 0.90 & 0.10 && 23.27 & 0.92 & 0.09 && 17.58 & 0.86 & 0.16 && 17.80 & 0.84 & 0.20\\
        Plenoxels~\cite{yu2021plenoxels} &  &  21.97 & 0.90 & 0.11 && 25.18 & 0.93 & 0.08 && 18.76 & 0.87 & 0.15 && 20.18 & 0.86 & 0.19\\
        InstantNGP~\cite{mueller2022instant} && 19.99 & 0.90 & 0.11 && 25.45 & 0.94 & 0.07 && 15.99 & 0.86 & 0.20 && 19.62 & 0.90 & 0.17\\
        \cline{1-1} \cline{3-5} \cline{7-9} \cline{11-13} \cline{15-17} 
        T-NeRF~\cite{pumarola2021d} &  &  23.82 & 0.90 & 0.15 && 30.19 & 0.96 & 0.13 && 31.24 & 0.97 & 0.02 && 32.01 & 0.97 & \textbf{0.03}\\
        D-NeRF~\cite{pumarola2021d} &  &  21.64 & 0.83 & 0.16 && 31.75 & 0.97 & 0.03 && 32.79 & 0.98 & 0.02 && 32.80 & \textbf{0.98} & \textbf{0.03}\\ 
        NR-NeRF~\cite{tretschk2021non} && 23.90 & 0.91 & 0.14 && 28.28 & 0.96 & 0.12 && 26.61 & 0.96 & 0.05 && 24.70 & 0.94 & 0.09\\
        TiNeuVox-S \cite{fang2022fast} &&  24.35 & 0.88 & 0.13 && 29.95 & 0.96 & 0.06 && 32.89 & 0.98 & 0.03 && 32.33 & 0.97 & 0.04\\
        TiNeuVox-B \cite{fang2022fast} && 25.02 & 0.92 & 0.07 && 32.70 & 0.98 & 0.03 && \textbf{35.43} & \textbf{0.99} & 0.02 && \textbf{34.23} & \textbf{0.98} & \textbf{0.03}\\
        NDVG~\cite{guo2022neural} && 25.04 & 0.94 & 0.05 && 32.62 & 0.98 & 0.03 && 33.22 & 0.98 & 0.03 && 31.25 & 0.97 & 0.04\\
        $K$-Planes-hybrid~\cite{kplanes_2023} && \textbf{25.48} & \textbf{0.95} & -- && 31.79 & 0.98 & -- && 33.72 & 0.98 & -- && 32.64 & \textbf{0.98} & --\\
        \textbf{MoNeRF} && 25.19 & 0.94 & \textbf{0.04} && \textbf{33.06} & \textbf{0.99} & \textbf{0.02} && 34.29 & \textbf{0.99} & \textbf{0.01} && 32.14 & \textbf{0.98} & \textbf{0.03}\\
        \hline
    \end{tabular}
    }
\end{table*}

\begin{figure*}[ht]
    \centering
    \includegraphics[width=\textwidth,keepaspectratio]{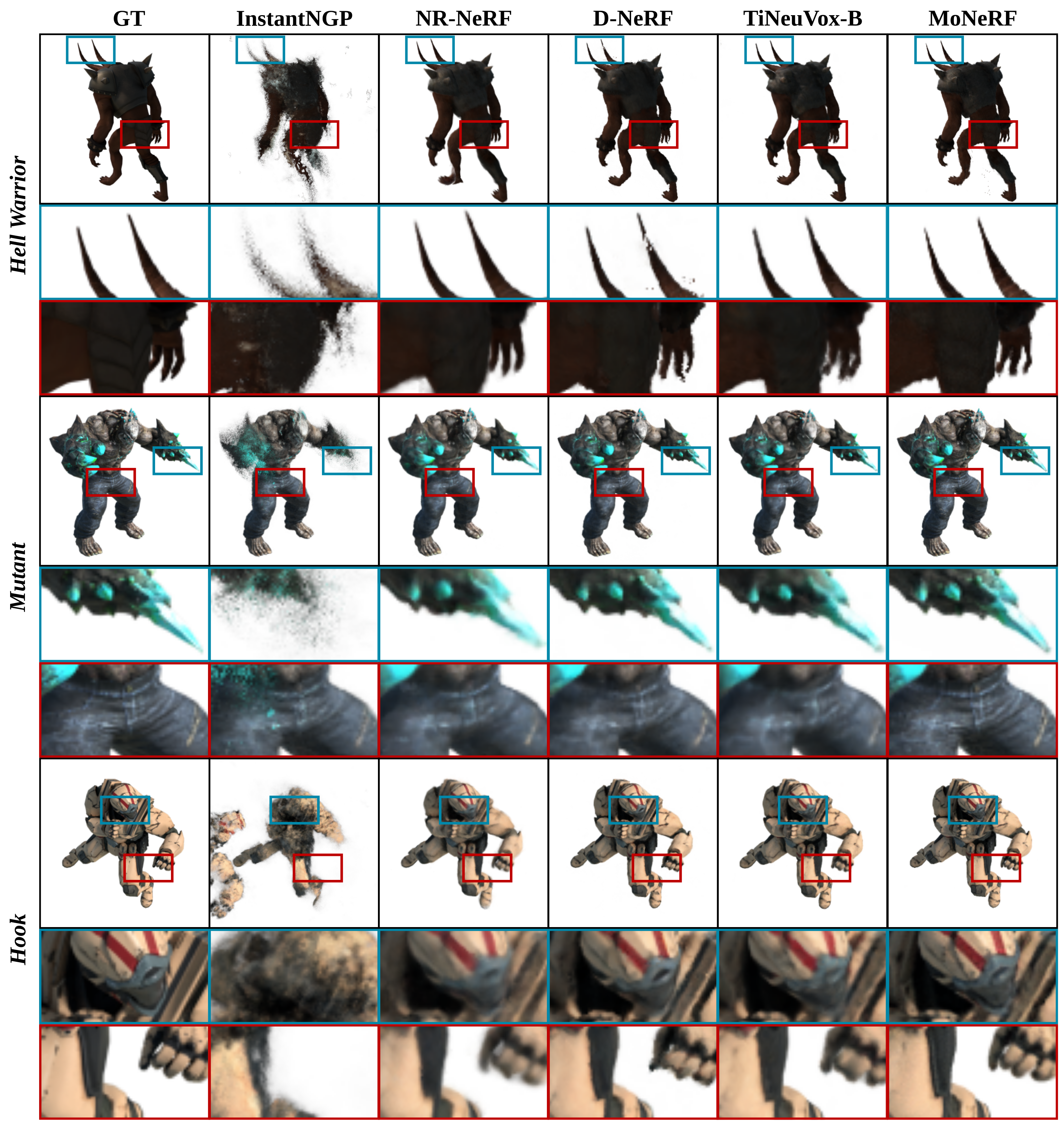}
    \label{fig:SUP-CompDNeRF-1-2} 
\end{figure*}

\begin{figure*}[ht]
    \centering
    \includegraphics[width=\textwidth,keepaspectratio]{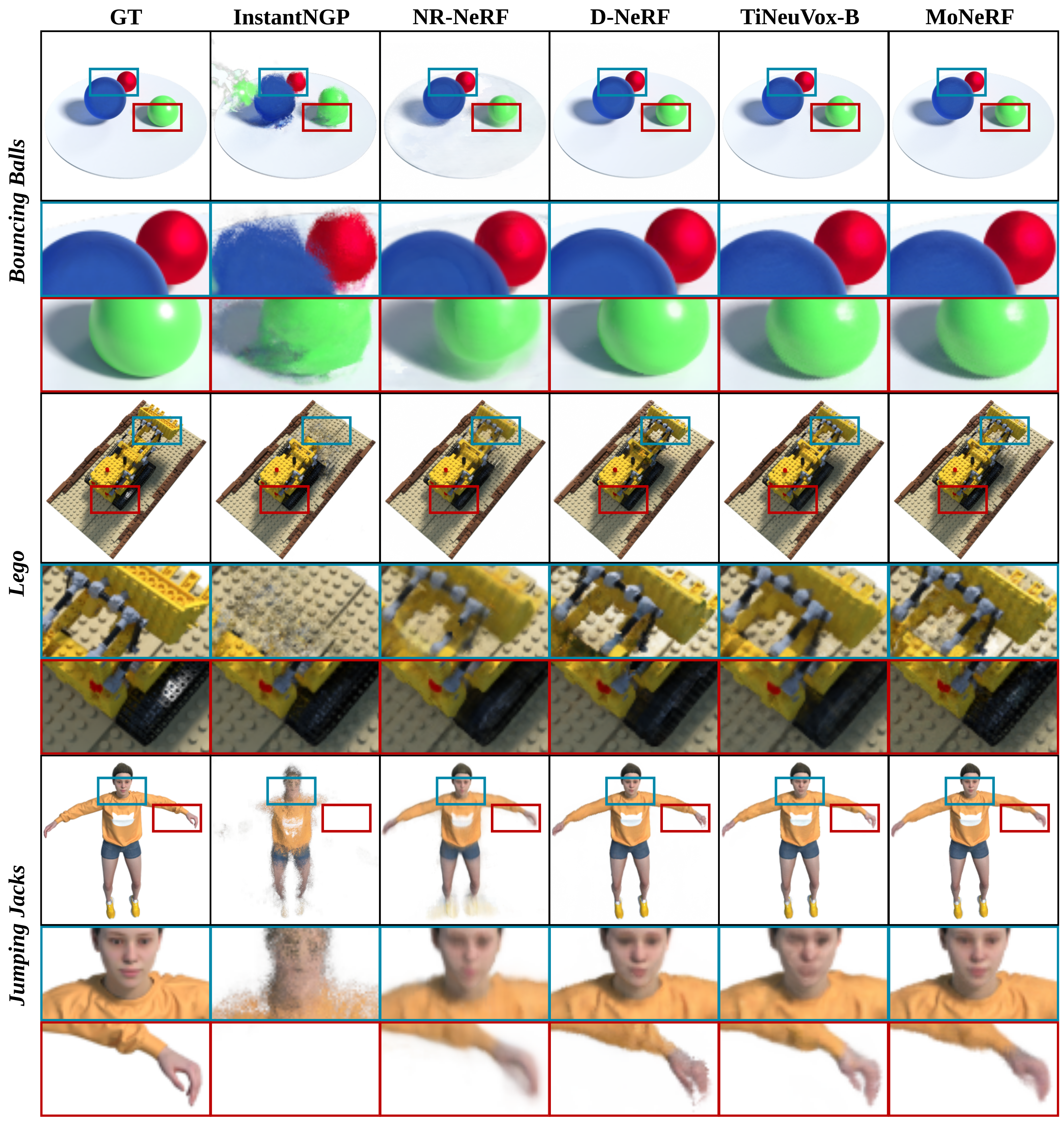} 
    \caption{
        \textbf{Qualitative comparisons on the D-NeRF~\cite{pumarola2021d} dataset}: Novel view synthesis on the remaining six scenes. 
    } 
    \label{fig:SUP-CompDNeRF-2-2} 
\end{figure*} 
\begin{table*}[ht]
    \caption{\textbf{Quantitative per-scene comparison on our MMVA dataset}. We report PSNR/SSIM (higher is better) and LPIPS (lower is better) on all twelve dynamic scenes of our dataset. The best results are highlighted in bold.}\label{tab:MetricsOurDB}
    \centering
    \small{
    \setlength\tabcolsep{2pt}
    \begin{tabular}{lcccccccccccccccc}
        \hline
         &  & \multicolumn{3}{c}{\textit{Archer}} &  & \multicolumn{3}{c}{\textit{Banner}} &  & \multicolumn{3}{c}{\textit{Bow}} &  & \multicolumn{3}{c}{\textit{Fusion}} \\
        \textbf{Method} &  & \textbf{PSNR $\uparrow$} & \textbf{SSIM $\uparrow$} & \textbf{LPIPS $\downarrow$} &  & \textbf{PSNR $\uparrow$} & \textbf{SSIM $\uparrow$} & \textbf{LPIPS $\downarrow$} &  & \textbf{PSNR $\uparrow$} & \textbf{SSIM $\uparrow$} & \textbf{LPIPS $\downarrow$} &  & \textbf{PSNR $\uparrow$} & \textbf{SSIM $\uparrow$} & \textbf{LPIPS $\downarrow$} \\ \cline{1-1} \cline{3-5} \cline{7-9} \cline{11-13} \cline{15-17} 
        InstantNGP~\cite{mueller2022instant} && 14.774 & 0.894 & 0.286 && 19.892 & 0.953 & 0.069 && 14.482 & 0.883 & 0.238 && 16.128 & 0.932 & 0.164 \\
        \cline{1-1} \cline{3-5} \cline{7-9} \cline{11-13} \cline{15-17} 
        D-NeRF~\cite{pumarola2021d} && 32.578 & 0.985 & 0.021 && 31.172 & 0.979 & 0.161 && 35.229 & 0.992 & 0.059 && 32.936 & 0.988 & 0.144\\
        NR-NeRF~\cite{tretschk2021non} && 30.604 & 0.983 & 0.022 && 31.877 & 0.986 & 0.029 && 30.408 & 0.987 & 0.023 && 30.993 & 0.989 & 0.018\\
        TiNeuVox-S~\cite{fang2022fast}&& 33.738 & 0.987 & 0.018 && 32.563 & 0.987 & 0.020 && 36.475 & 0.995 & 0.011 && \textbf{34.194} & 0.992 & 0.014\\
        TiNeuVox-B~\cite{fang2022fast}&& 34.151 & 0.989 & 0.016 && 32.515 & 0.987 & 0.021 && 36.049 & 0.994 & 0.013 && 34.128 & \textbf{0.993} & 0.015\\
        \textbf{MoNeRF} && \textbf{34.535} &\textbf{ 0.991} &\textbf{ 0.014} && \textbf{33.651} & \textbf{0.991} & \textbf{0.013} && \textbf{37.175} & \textbf{0.996} & \textbf{0.007} && 33.869 & \textbf{0.993} & \textbf{0.013}\\
        \hline
        &  & \multicolumn{3}{c}{} &  & \multicolumn{3}{c}{} &  & \multicolumn{3}{c}{} &  & \multicolumn{3}{c}{} \\
        &  & \multicolumn{3}{c}{\textit{Jacket}} &  & \multicolumn{3}{c}{\textit{Kimono}} &  & \multicolumn{3}{c}{\textit{PlushDog}} &  & \multicolumn{3}{c}{\textit{Sari}} \\
        \textbf{Method} &  & \textbf{PSNR $\uparrow$} & \textbf{SSIM $\uparrow$} & \textbf{LPIPS $\downarrow$} &  & \textbf{PSNR $\uparrow$} & \textbf{SSIM $\uparrow$} & \textbf{LPIPS $\downarrow$} &  & \textbf{PSNR $\uparrow$} & \textbf{SSIM $\uparrow$} & \textbf{LPIPS $\downarrow$} &  & \textbf{PSNR $\uparrow$} & \textbf{SSIM $\uparrow$} & \textbf{LPIPS $\downarrow$} \\ \cline{1-1} \cline{3-5} \cline{7-9} \cline{11-13} \cline{15-17} 
        InstantNGP~\cite{mueller2022instant} && 22.945 & 0.977 & 0.023 && 15.133 & 0.910 & 0.306 && 15.696 & 0.889 & 0.298 && 14.303 & 0.896 & 0.362\\
        \cline{1-1} \cline{3-5} \cline{7-9} \cline{11-13} \cline{15-17} 
        D-NeRF~\cite{pumarola2021d} && 29.570 & 0.989 & 0.017 && 30.437 & 0.980 & 0.020 && 29.797 & 0.985 & 0.022 && 32.849 & 0.982 & 0.070\\ 
        NR-NeRF~\cite{tretschk2021non} && 30.767 & 0.990 & 0.015 && 29.035 & 0.977 & 0.024 && 29.520 & 0.983 & 0.029 && 30.054 & 0.980 & 0.026\\
        TiNeuVox-S~\cite{fang2022fast}&& 36.584 & 0.995 & 0.009 && 30.849 & 0.981 & 0.020 && 35.200 & 0.993 & 0.012 && 32.725 & 0.984 & 0.020\\
        TiNeuVox-B~\cite{fang2022fast}&&36.413 & 0.995 & 0.008 && \textbf{31.411} & 0.982 & \textbf{0.019} && 34.801 & 0.993 & 0.012 && \textbf{33.687} & 0.985 & 0.019\\
        \textbf{MoNeRF} && \textbf{37.925} & \textbf{0.997} & \textbf{0.006} && 31.378 & \textbf{0.984} & \textbf{0.019} && \textbf{35.654} & \textbf{0.994} & \textbf{0.009} && 33.446 & \textbf{0.988} & \textbf{0.015}\\
        \hline
        &  & \multicolumn{3}{c}{} &  & \multicolumn{3}{c}{} &  & \multicolumn{3}{c}{} &  & \multicolumn{3}{c}{} \\
        &  & \multicolumn{3}{c}{\textit{Scissors}} &  & \multicolumn{3}{c}{\textit{Squat}} &  & \multicolumn{3}{c}{\textit{Stability}} &  & \multicolumn{3}{c}{\textit{Umbrella}} \\
        \textbf{Method} &  & \textbf{PSNR $\uparrow$} & \textbf{SSIM $\uparrow$} & \textbf{LPIPS $\downarrow$} &  & \textbf{PSNR $\uparrow$} & \textbf{SSIM $\uparrow$} & \textbf{LPIPS $\downarrow$} &  & \textbf{PSNR $\uparrow$} & \textbf{SSIM $\uparrow$} & \textbf{LPIPS $\downarrow$} &  & \textbf{PSNR $\uparrow$} & \textbf{SSIM $\uparrow$} & \textbf{LPIPS $\downarrow$} \\ \cline{1-1} \cline{3-5} \cline{7-9} \cline{11-13} \cline{15-17} 
        InstantNGP~\cite{mueller2022instant} && 15.524 & 0.927 & 0.197 && 13.730 & 0.896 & 0.533 && 16.630 & 0.960 & 0.091 && 22.286 & 0.967 & 0.202\\
        \cline{1-1} \cline{3-5} \cline{7-9} \cline{11-13} \cline{15-17} 
        D-NeRF~\cite{pumarola2021d} &&  35.315 & 0.993 & 0.015 && 32.882 & 0.988 & 0.016 && 32.328 & 0.991 & 0.066 && 35.472 & 0.992 & 0.044\\ 
        NR-NeRF~\cite{tretschk2021non} && 31.625 & 0.990 & 0.015 && 32.036 & 0.988 & 0.016 && 31.681 & 0.990 & 0.020 && 33.747 & 0.992 & 0.010\\
        TiNeuVox-S~\cite{fang2022fast}&& 36.411 & 0.994 & 0.012 && 34.897 & 0.991 & 0.013 && 33.383 & 0.993 & 0.009 && 35.918 & 0.994 & 0.008\\
        TiNeuVox-b~\cite{fang2022fast}&& 35.826 & 0.994 & 0.012 && 34.716 & 0.991 & \textbf{0.012} && 33.970 & 0.993 & 0.010 && 35.444 & 0.993 & 0.008 \\
        \textbf{MoNeRF} && \textbf{36.718} & \textbf{0.996}  & \textbf{0.008}  && \textbf{34.950}  & \textbf{0.993}  & \textbf{0.012}  && \textbf{35.113}  & \textbf{0.995}  & \textbf{0.007}  && \textbf{36.534}  & \textbf{0.995}  & \textbf{0.006}\\
        \hline
    \end{tabular}
    }
\end{table*}
\begin{figure*}[ht]
    \centering
    \includegraphics[width=\textwidth,keepaspectratio]{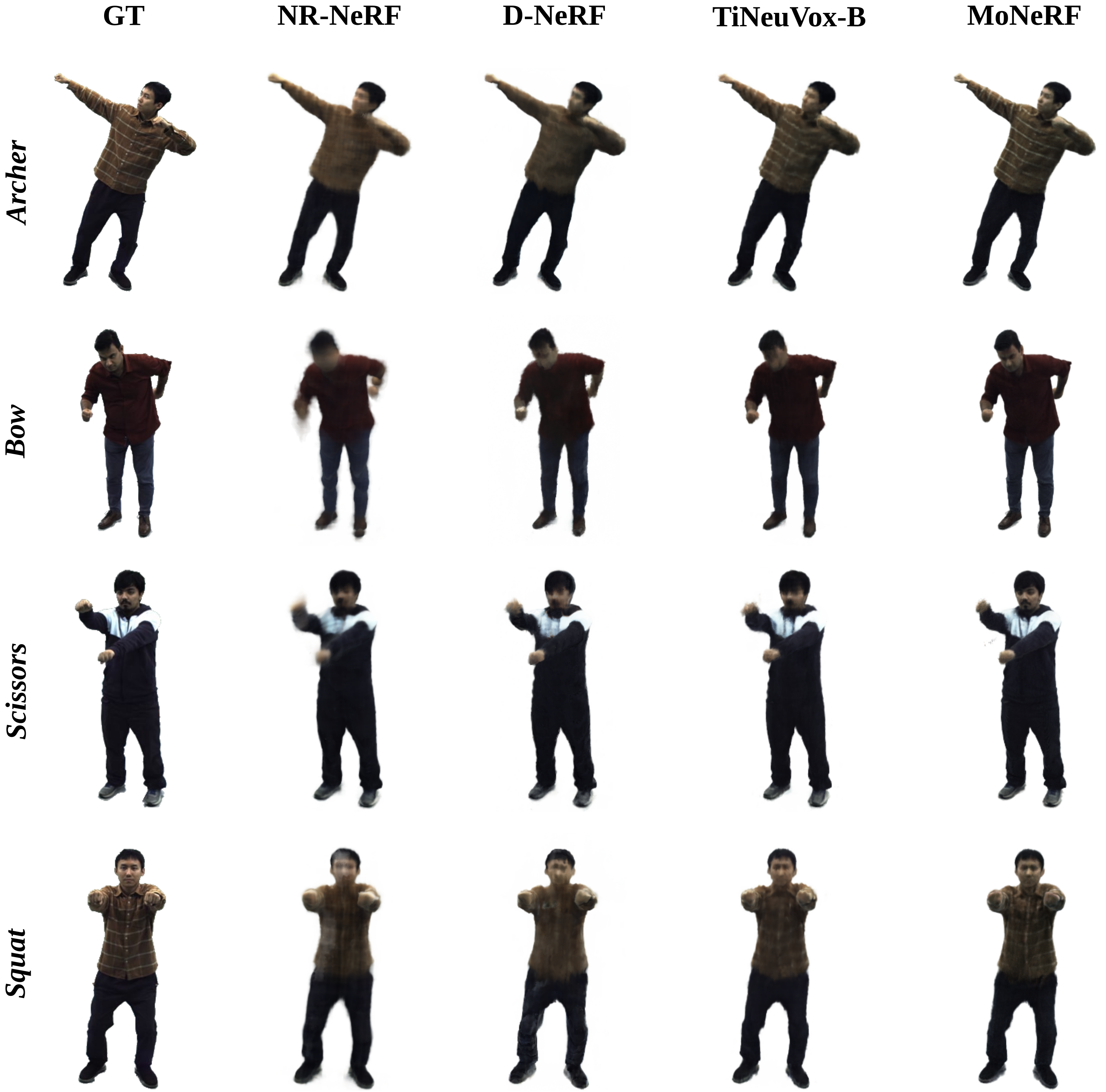}
    \caption{
        \textbf{Further qualitative comparisons on our MMVA dataset}: Novel view synthesis on the remaining four scenes. 
    } 
    \label{fig:SUP-CompMMVA} 
\end{figure*} 
In \cref{tab:MetricsDNeRF} and \cref{tab:MetricsOurDB}, we supplement our experiments with detailed per-sequence quantitative analysis for the D-NeRF and MMVA datasets, respectively. 
We also provide more qualitative comparisons in \cref{fig:SUP-CompDNeRF-2-2} and \cref{fig:SUP-CompMMVA}, showing all remaining dataset sequences not contained in the main document.
Please see our supplemental video for in depth dynamic comparisons.
\vfill

\putbib

\end{bibunit}

\end{document}